\documentclass[5p,times]{elsarticle}

\usepackage{hyperref, lineno}
\usepackage{amssymb}
\usepackage{subcaption}
\usepackage{array}
\usepackage{colortbl}
\usepackage{textcomp}
\usepackage{makecell}
\usepackage{listings}
\usepackage{textcomp }
\usepackage{multirow}
\usepackage{hyperref}
\usepackage[normalem]{ulem}
\usepackage{amsmath}
\usepackage{xurl}
\usepackage{algorithm}
\usepackage[table,xcdraw]{xcolor}
\usepackage[noend]{algpseudocode}
\usepackage{graphicx}
\usepackage{tabularx}
\usepackage{manfnt}
\useunder{\uline}{\ul}{}
\newcommand{\ans}[1]{{\color{black}#1}}
\newcommand{\ed}[1]{{\color{black}#1}}

\modulolinenumbers[5]

\journal{Algorithms}

\begin{document} \sloppy

\begin{frontmatter}

\title{Improvement of Computational Performance of Evolutionary AutoML in a Heterogeneous Environment}


\author{Nikolay O. Nikitin}
\ead{nnikitin@itmo.ru}

\author{Sergey Teryoshkin}

\author{Valerii Pokrovskii}

\author{Sergey Pakulin}

\author{Denis Nasonov}

\address{ITMO University, Saint-Petersburg, Russia}

\begin{abstract}
Resource-intensive computations are a major factor that limits the effectiveness of automated machine learning solutions. In the paper, we propose a modular approach that can be used to increase the quality of evolutionary optimization for modelling pipelines with a graph-based structure. It consists of several stages - parallelization, caching and evaluation. Heterogeneous and remote resources can be involved in the evaluation stage. The conducted experiments confirm the correctness and effectiveness of the proposed approach. The implemented algorithms are available as a part of the open-source framework FEDOT.
\end{abstract}


\end{frontmatter}


\section{Introduction}

Nowadays, automated machine learning (AutoML) is widely used in science, and industry \cite{li2021automl, singh2022automated}. The major problem of solving real-world tasks with AutoML is the high computational cost of the search for an optimal modelling pipeline. During the evaluation of the candidate pipelines' quality, many machine learning models are trained. This task is very resource-intensive, so \ed{it can take a considerable amount of time to} achieve the appropriate result. It can be considered a bottleneck for any existing AutoML \ed{solution}. This issue raises various problems from different fields: from integration of AutoML to business processes \cite{santu2020automl} to carbon emission and sustainability concerns\cite{tornede2021towards}.

There are many approaches for improving computational performance that are used in state-of-the-art (SOTA) AutoML solutions \cite{he2021automl}. First of all, almost all solutions support parallel execution. Some of them also support caching of evaluated candidates \cite{olson2016tpot}. Also, the graphics processing unit can be used to reduce the training time \cite{ledell2020h2o}. 

 There is a variety of open-source tools that could improve the efficiency of certain steps of machine learning pipelines. For instance, involving various MLOps tools like MLFlow \cite{zaharia2018accelerating}, task-specific databases \cite{visheratin2020peregreen} and scaling tools like Ray \cite{moritz2018ray} allows the effectiveness of ML applications to be notably increased.

However, the optimal design of the computational strategy depends on the infrastructure and the underlying AutoML algorithm. The SOTA AutoML solutions are based on different optimization methods: random search, Bayesian optimization, genetic algorithms, and meta-learning \cite{he2021automl, cohen2022learning}. The structural patterns used in modelling pipelines can also be different: linear pipelines or ensembling techniques (stacking, blending, and boosting) \cite{zoller2019benchmark}. The most complicated case is a composite pipeline represented as a directed acyclic graph \cite{nikitin2022automated}. At the same time, there is no ready-to-use solution for improving the computational performance of an automated open-ended search for pipelines in the composite AI field.

In the paper, we want to propose a \ans{adaptive approach} to reduce the computational cost of AutoML for composite pipelines. Several techniques are implemented: pipeline caching, parallelization of the fitness function evaluation, computation with hybrid (GPU and CPU) systems, and integration with remote distributed systems. 

\ans{This approach differs from existing solutions since it can be configured for automated machine learning in various computational environments (including distributed and heterogeneous). Also, the caching procedure can be effectively used for various pipeline designs (linear, weighted ensembles, multi-layer ensembles, etc). 

To confirm the effectiveness of the proposed approach in empirical way, we conducted a set of numerical experiments using set of open datasets of various sizes (described in Table~\ref{tab_datas}). The results presented in Section~\ref{sec_exp} allow us to conclude that a larger number of pipelines can be evaluated and better quality metrics can be achieved by AutoML using this approach.} The software implementation is available in the open-source AutoML framework FEDOT.

The paper is organized as follows: Section~\ref{sec_related} describes the computational strategies used in state-of-the-art AutoML tools. \ans{Section~\ref{sec_problem} provides the problem statement for AutoML performance improvement}. Section~\ref{sec_approach} proposes a set of novel improvements for the composite evolutionary AutoML. Section~\ref{sec_software} describes the software implementation of these techniques in an open-source framework. Section~\ref{sec_exp} provides the experimental evaluation of the proposed techniques for different case studies. Finally, Sec.~\ref{sec_conc} provides an analysis of the obtained results and possible extensions of the research.

\section{\ans{Related Works}}
\label{sec_related}

There are dozens of open-source AutoML solutions that can be used for designing modelling pipelines. The first frameworks that became well-known are H2O \citep{ledell2020h2o}, TPOT \citep{le2020scaling} and Auto-sklearn \citep{feurer2020auto}. As more novel AutoML solutions, AutoGluon \citep{erickson2020autogluon} and LAMA \citep{vakhrushev2021lightautoml} can be noted. Also, there are a lot of other AutoML tools with various specific features \cite{majidi2022empirical}.

There are different strategies for performance improvement used in the noted frameworks. In the TPOT framework, pipeline caching is implemented \cite{le2020scaling}. TPOT-SH \cite{parmentier2019tpot} uses the concept of Successive Halving to explore the search space faster, especially for larger datasets. Various techniques are used to evaluate the pipelines on different subsets of training data (e.g. layering \cite{gijsbers2018layered}). 

\ed{A} widely-used parallelization tool is the joblib library implemented in Python. However, there are more advanced frameworks for parallelization that can be noted. For example, Ray \cite{moritz2018ray} can be used to scale AI and Python applications in distributed environments. It provides various instruments for distributed data preprocessing, distributed training of ML models and scalable hyperparameter tuning.

Improving the computational performance for evolutionary algorithms outside AutoML is also discussed in the literature. As an example, parallel GPU-based evaluation of the fitness function can be used \cite{orzechowski2019strategies} to solve the expensive problems related to big data \cite{orzechowski2019mining}. There are various techniques that improve the performance of evolutionary algorithms in concurrent mode \cite{guervos2019improving}. The tensor-based computational model can be used to achieve cross-platform hardware acceleration \cite{klosko2022high}. Also, platform-specific open-source solutions are presented in \ed{this} field (e.g. scikit-learn-intelex \footnote{https://github.com/intel/scikit-learn-intelex}).

One of the widely used techniques to avoid fitness evaluation bottlenecks in evolutionary algorithms is caching \cite{przewozniczek2021fitness}. Final values of the fitness evaluation can be cached \cite{kratica1999improving} as well as partial results \cite{santos2001effective, kim2009effective, karatsiolis2014implementing}.

Moreover, a number of solutions exist that can perform remote/distributed training (e.g. Auto-sklearn, H2O, TPOT, LAMA). These AutoML frameworks use different frameworks for distributed computing. Autosklearn and TPOT use Dask\footnote{\url{https://dask.org}}, LAMA uses Apache Spark\footnote{\url{https://spark.apache.org}}. H2O uses its own Apache Spark modification called Sparkling Water\footnote{\url{https://h2o.ai/products/h2o-sparkling-water}}. \ed{Distributed computing frameworks allow the processing of large datasets spread over the nodes of a cluster system.}

We can conclude that there is a large number of techniques and solutions that can reduce the resource consumption for AutoML and EA. However, there is still no well-developed approach that can be used to identify graph-based pipelines in the heterogeneous computational environment in composite AI problems. For this reason, we decided to formulate the problem statement specific to composite AutoML and propose possible solutions.

\section{\ans{Problem Statement}}
\label{sec_problem}

We want to design multi-task and multi-modal pipelines for various tasks using a single flexible instrument. Consequently, it becomes necessary to implement the framework's architecture more abstractly to separate the pipeline search process from the top-level API. The modelling pipeline is represented as a directed acyclic graph in this case. Each node (modelling or data transformation operation) is described by the operation's name and set of hyper-parameters. If necessary, different data sources (tables, time series, images, texts) can be involved in the pipeline. Also, metadata is attached to the data flow, making it possible to change the task several times during the pipeline evaluation (e.g., solve a classification task and then - a regression task). 

The drawback of this approach is the increased search space that should be explored during the optimization. In automated modelling, we want to control the balance between open-endedness \citep{packard2019open} and local search. \ans{The simplest way is to apply the of direct constraints (e.g. limit to the pipeline size). Also, it can be more effective to apply the regularization and sensitivity analysis procedures \cite{nikitin2022automated} and adaptive optimisation strategies to control the convergence of optimisation.} At the same time, avoiding over-complicated pipelines and reckless spending of a limited time budget is also essential. It makes the effectiveness of the computational part even more critical for open-ended AutoML.

It pushes us to compromise between pipeline complexity and training time. However, if we can improve computing efficiency, the framework will probably be able to build more complicated models with higher quality while consuming the same training time. There are many approaches to improving evolutionary algorithms' computing performance, such as parallelization, caching, etc. These approaches can be divided into single-machine optimizations and horizontal scaling techniques. Single-machine optimizations aim to improve computing performance only on the machines performing the computations. Horizontal scaling allows involving additional servers to speed up computing. Both techniques can be used separately or combined.

Evolutionary algorithms' computational time mainly depends on the population size and the number of generations. Increasing population size leads to an increased probability of getting better individuals. A more significant number of generations means more attempts to grow better individuals based on the best previous generation.

From a computational point of view, we have several iterations, each of which requires the results from the previous iteration. So, it is complicated to scale computations over the iterations, and the total computation time is Equation~\ref{eq_total_training_time}.

\begin{equation}
\label{eq_total_training_time}
T_{total}=\sum_{i=1}^{n} T_{i}
\end{equation}

\ans{where \(n\) is the number of iterations and  \(T_{i}\) - the computation time of generation \(i\). Inside one generation, all individuals are processed independently from one another, allowing us to scale these computations according to the available computational resources. The population training time can be estimated with Equation~\ref{eq_population_training_time}.}

\begin{equation}
\label{eq_population_training_time}
T_{total} =\sum_{i=1}^{n} \underset {d_{j}^{i}\in D_{i}}{\operatorname*{argmax}}(\underset {r_{j}\in R_{i}/\{r_{j-1},..,r_{1}\}}{\operatorname*{argmin}(\tau(d_{j}^{i},} \omega(\scriptstyle { D^{i-1},..,D^{1}}\displaystyle),r_{j})) 
\end{equation}
where \(D_{i}\) is the set of individuals in the population \(i\) and \(d_{j}^{i}\) is \ed{individual \(j\)}, \(R_{i}\) is the set of available resources on iteration \(i\), \(\omega\) is a cache function with pre-calculated elements on iterations \(i-1, i-2, ... , 1\), and \(\tau\) is a function that returns the calculation time considering caching and evaluation of individuals \(D_{i}\).

Due to the \ans{Equations}~\ref{eq_total_training_time} and \ref{eq_population_training_time}, we should increase the population size as much as possible. It allows to speed up the algorithm convergence and improve its result using horizontal scaling. For this purpose, we can use both remote computing on production servers and distributed computing using homogeneous and heterogeneous computing clusters.

Remote computing allows model training to be delegated to a remote infrastructure. This approach is justified if the local machine computational resources are insufficient to train a set of models in a reasonable amount of time. Remote computing may take place on a dedicated computing server or a cluster of servers that \ed{accepts} tasks to train models using REST API, RPC or message queues.

\begin{figure*}[h!] 
    \centering
    \includegraphics[width=16.3cm,height=7.8cm]{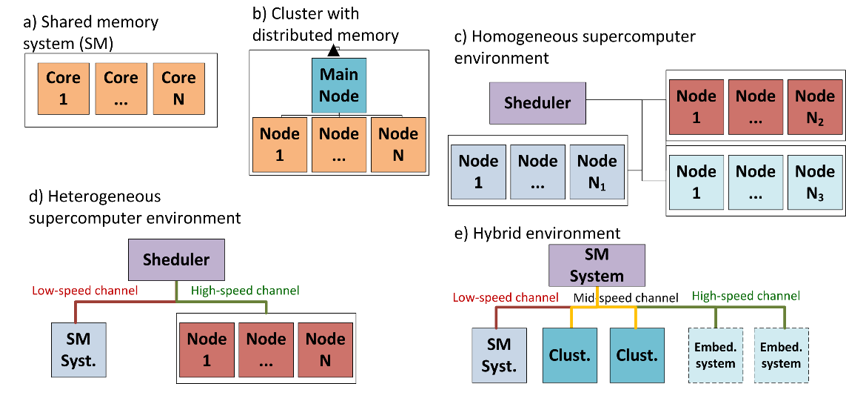}
    \caption{Different types of computational infrastructures that can be used in AutoML: (a) shared memory (SM) system (b) multi-node cluster with distributed memory (c) complex homogeneous supercomputer system with spatially distributed infrastructure (d) supercomputer system with heterogeneous distributed infrastructure.}
    \label{fig_infras}
\end{figure*}

\ans{The main challenge in the investigated problem is to propose the performance improvement strategy for AutoML that is adaptive to various types of computational infrastructures. In Figure~\ref{fig_infras}, five classes are noted: shared memory system, multi-node cluster with distributed memory, complex homogeneous and heterogeneous supercomputer environments, and hybrid systems. The system with structure (a) can execute parallel tasks in a straightforward way. In the systems (b)-(e), the remote nodes are involved (homogeneous and heterogeneous). For the system (e), the structure is hybrid since various remote nodes have different computational performance and connections overheads. For this reason, the adaptability of the computational strategy is especially important.}

\ans{There are various ways can be uses to adapt the computational strategy to specific infrastructure.} For example, the empirical performance models \cite{kalyuzhnaya2020towards} can be used to choose the optimal infrastructure for evaluating specific pipelines. Simple pipelines with low fitting time can be assigned to low-performance computational nodes. Otherwise, complicated pipelines with high fitting time can be assigned to high-performance nodes. It makes it necessary to develop a modular approach that effectively utilizes all available resources.

\ans{Our main motivation is to develop an approach that can be used at the computational layer of AutoML. It should be possible to adapt this layer to the specified  infrastructure (local or remote) in a frame of the same AutoML approach. This solution should be high-level, modular and flexible to allow integrating it with different AutoML tools. Also, it should support the different types of pipelines (from simplest linear pipeline to the multi-level ensembles).}

\section{\ans{Proposed Improvements}}
\label{sec_approach}

\begin{figure}[ht!] 
    \centering
    \includegraphics[width=8cm]{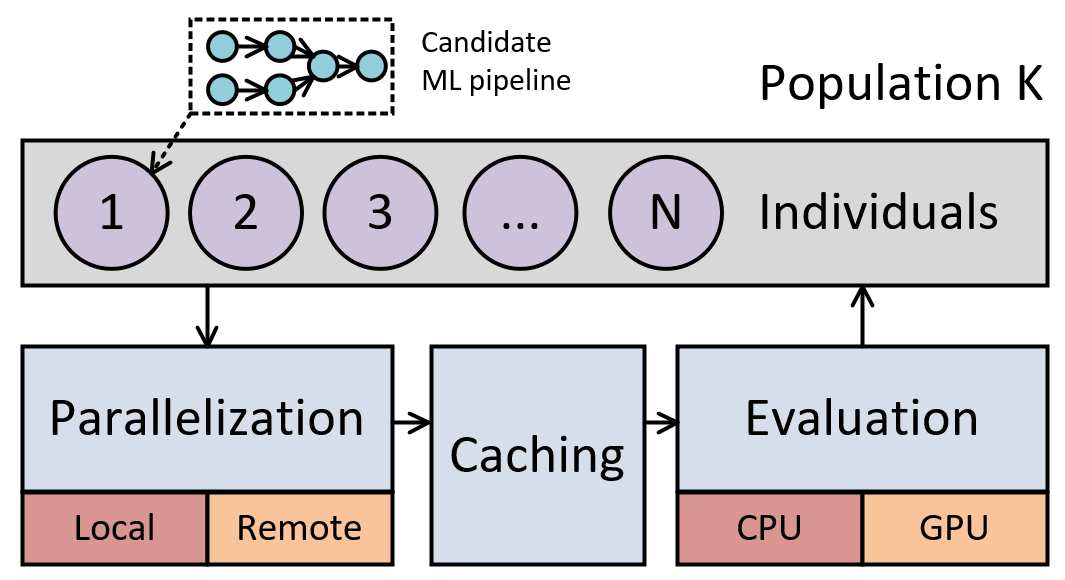}
    \caption{Workflow of the proposed approach for the improvement of computational performance for composite AutoML}
    \label{fig_all}
\end{figure}

This section is devoted to various aspects of the proposed approach for improving the computational performance of evolutionary AutoML. The high-level scheme of the approach is presented in Figure~\ref{fig_all}. Four main aspects are considered: (1) parallelization of the fitness function evaluation; (2) partial caching of evaluated individuals; (3) combining CPU and GPU to accelerate the processing of individuals (4) integration with remote infrastructure for a complex task. Algorithmic-based improvements (e.g. surrogate-assisted optimization) are not considered here.

The \ed{detailed} implementation of the proposed approach is described in Alg.~\ref{alg_parall}. In this notation, \textit{graph} represents the structure of \ed{the} composite pipeline. The details \ed{of} the evolutionary optimisation are hidden to make the proposed improvements more clear.

\begin{algorithm}[ht!]
\caption{\ans{High-level pseudocode of the evaluation dispatching algorithm implemented in the proposed approach. Parallelization, caching and evaluation stages are demonstrated for processing one generation of the evolutionary algorithm.}}\begin{algorithmic}[1]
\Procedure{ProcessPopulation}{}
\State \underline{Input:}  \newline
 $inds \text{ (set of non-evaluated individuals), } \newline
 objective \text{ (objective function that calculates the fitness}\newline \text{of an individual), }\newline
 n \text{ (number of parallel jobs)} \newline
 timer \text{ (timer-like object)} \newline
 infrastructure \text{ (description of setup)}$
 \State \underline{Output:} $evaluated\_inds$
 \State {\textbf{do in parallel(n)}}
    \If{\Call{timer.enough\_time}{ }}
        \State $graph \gets inds[i].graph$ \Comment{get structure of each ind.}
        \If{\Call{infrastructure.is\_remote}{ }}
  \State $cache \gets DistributedCache()$
  \State \textbf{sync} cache \Comment{sync cache database}
  \State $task\_id \gets \Call{create\_task}{graph}$
  \State \textbf{wait} $task\_id$
  \State $inds[i].fitness \gets \Call{request\_result}{graph}$
        \Else 
  \State \textbf{prepare} $graph$ \Comment{assign CPU and GPU to nodes}
  \State $cache \gets LocalCache()$
  \State \textbf{load} cache \Comment{init cache database}
  \If{\Call{cache.exists(graph)}{ }}
      \State \Call{fit\_from\_cache}{graph}
      \State $inds[i].fitness \gets \Call{obj}{graph, cache}$
  \EndIf 
  \State \Call{fit}{graph} \Comment{Fit nodes that are not in cache}
  \State \textbf{save} cache \Comment{preserve updated cache}
  \EndIf 
  \If{not $inds[i].is\_valid$}
      \State \textbf{delete} $inds[i]$  \Comment{for unsuccessful evaluation}
        \EndIf 
    \Else
        \State \textbf{delete} $inds[i]$ \Comment{not enough time, skipping}
    \EndIf 
    
 \State \textbf{return} $inds$ \Comment{candidates for selection}
\EndProcedure
\end{algorithmic}
\label{alg_parall}
\end{algorithm}

\subsection{Parallelization}

\ed{Parallelizing evolutionary algorithms is not a novel idea.} There are a lot of papers and open-source solutions devoted to this problem. However, parallelization in AutoML has its specifics. For example, various computationally efficient strategies of parallel evolution can be used \cite{kalyuzhnaya2020towards}.

We are considering an evolutionary algorithm for searching for the best solution in the space of pipelines that can be represented as directed acyclic graphs. The classic approach to parallelizing evolutionary optimization is evaluating \ed{all individuals} in the population concurrently \cite{kalyuzhnaya2020towards}. It works because of the nature of the evolutionary algorithm. There are no dependencies between individuals in a generation. Other approaches suggest dividing populations into isolated parts \cite{DaSilva2010} or using co-evolutionary algorithms to divide tasks into subtasks \cite{GaoY2016}. 

 \ans{\ed{The proposed} algorithm considers the maximum evaluation time length for each pipeline evaluation to resolve the possible evaluation time anomalies caused by the stochastic nature of data-driven model training. If the training process does not converge at least in one cross-validation fold, the time required for corresponding fitness evaluation can be increased significantly. So, the individuals that spend excess time on evaluation are skipped to preserve the overall performance of the evolutionary optimizer.}

\subsection{Caching}

The existing caching approaches are aimed at preserving and reusing fitted pipelines \cite{olson2016tpot}. However, separate nodes of composite pipelines can be cached individually \cite{nikitin2022automated}. It makes it possible to reuse the fitted models and reduce \ed{the fitness function's evaluation time}. The optimizer can share the in-memory cache across the populations, and individuals \cite{kalyuzhnaya2020towards}. However, it raises the problems of memory consumption.

After analyzing existing solutions, we focused on the relational database approach for pipeline caching. More specifically, the sqlite3 library was used to implement it. \ed{First of all}, it provides only one output file, which is not guaranteed for non-relational databases - e.g. shelve. Secondly, all concurrent save-load operations can be fully processed during the parallel evaluation of the fitness functions without direct usage of synchronization primitives, atomic variables and other instruments necessary for simultaneous access to data. 

\begin{figure}[h!] 
    \centering
    \includegraphics[width=8cm]{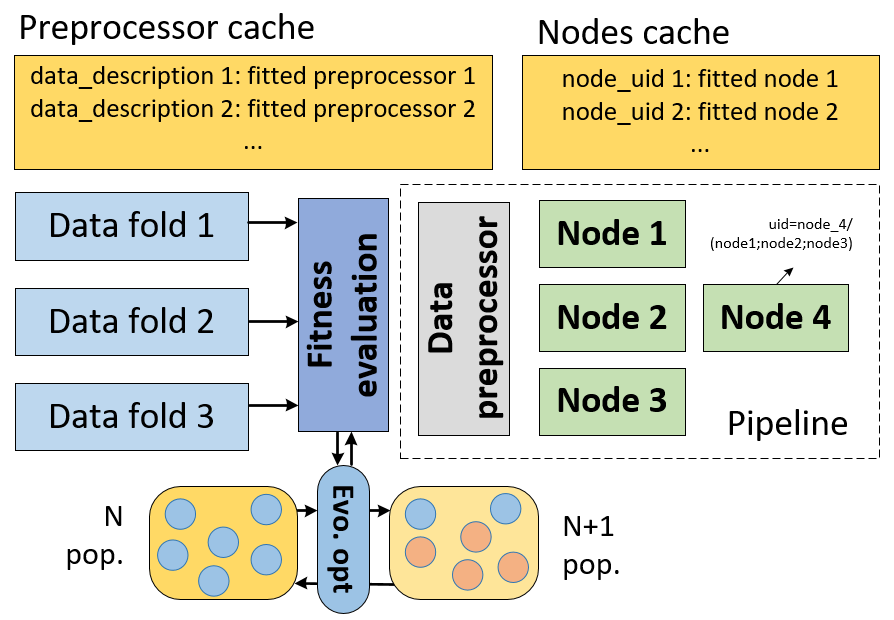}
    \caption{\ans{Interaction between operations' cache and the modelling pipeline that should be fitted}}
    \label{fig_cache_main}
\end{figure}

\ans{Finally, this approach allows extracting several operations simultaneously, which helps to improve the overall performance of caching. Also, the set of operations can be saved to the database taking into account the existence of cache items with the same primary key.

The caching procedure for the multi-layer ensemble pipelines should take into account that the cached model/operations are suitable only for the specific configuration of previous nodes and edges in the modelling pipelines. So, the key contains the recursive description of the structure of previous nodes and edges. Also, the identifier of cross-validation fold is specified. The following notation is used: \textit{(/[node\_name]\_[hparams];)/[node\_name]\_[hparams]..."}, where / denotes the beginning of the node name and round brackets represent the nested edges. The caching details are presented in Figure~\ref{fig_cache_main}.}

\subsection{GPU}

Evaluating ML models with \ed{GPUs} is a well-developed feature in many solutions. For example, the RAPIDS library \cite{rapids} contains the CuML module that allows training classification, regression and clustering models with \ed{GPUs}. To adapt this solution to composite pipelines, we should consider a setup in which only a part of the nodes can be evaluated with \ed{GPUs}. In this situation, the pipeline should be fitted in a heterogeneous way. 

The proposed approach makes it possible to use both \ed{CPUs} and \ed{GPUs} for fitting by separating the ML model type and its implementation. The same model (e.g. random forest) can have several implementations (CPU-based and GPU-based).

\begin{figure}[h!] 
    \centering
    \includegraphics[width=8cm]{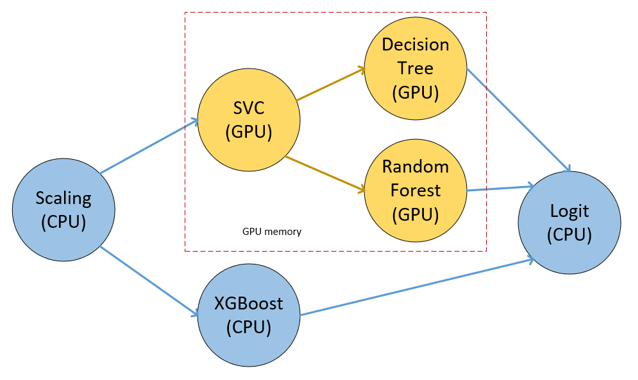}
    \caption{The structure of a pipeline that can be evaluated in a heterogeneous (CPU - blue and GPU - yellow) way}
    \label{fig_hetero_pipeline}
\end{figure}

 Figure~\ref{fig_hetero_pipeline} shows an example of a computationally heterogeneous composite model structure. Data transfer between the GPU-based nodes (yellow) is performed within the video memory, and the models themselves in the nodes are trained on graphics processing units (GPUs). Other nodes are executed on CPUs.
 
Due to the multiple software limitations set by RAPIDS libraries (CUDA-compatible GPU driver, restricted set of supported operation systems), it is practical to conduct the computations within Docker-based containers.

\subsection{Remote evaluation}

Remote evaluation can be integrated into the evolutionary optimiser in various ways. Both dataset folds and population parts can be distributed across several \ans{computational} nodes to satisfy time or memory limits. Since evolutionary algorithms do not always require processing large datasets, we have focused on the parallelism aspect of remote computing. The proposed implementation relies on Kubernetes. The REST API service inside the Kubernetes cluster is used to run computations via HTTP requests. The client implements a wrapper for requests. 

During the population training, the evaluator uses the client's methods to process individuals on the Kubernetes cluster. Then, after starting processing \ed{all individuals}, the evaluator waits for computations to be completed via client methods. The run request contains the container image, resources limit for the container, mount paths and model parameters. The REST API service creates the requested container and keeps monitoring it. The client uses requests to the REST API service to get actual containers' statuses to see if it is still running, completed or failed.

\begin{figure}[h!] 
    \centering
    \includegraphics[width=8cm]{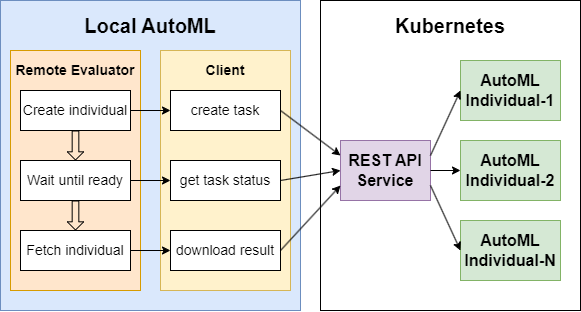}
    \caption{Communication between AutoML and remote cluster}
    \label{fig_communication_scheme}
\end{figure}

Finally, the client downloads the fitted pipeline when the training is completed. We wrap the result into a compressed archive to reduce the amount of data transferred over the network. Then, the files are sent to the client. This process scheme is presented in Figure~\ref{fig_communication_scheme}.

\ed{This way}, we can divide the population training process into three stages: (1) requests to evaluate individuals, (2) computing and waiting for the completion and (3) fetching the results.

\section{\ans{Software Implementation}}
\label{sec_software}

The proposed approach can be used as a part of the architecture that includes: 
\begin{itemize}

\item The model repository block, which provides storage and selection of various implementations of predictive models and data processing blocks. One model can contain several implementations (e.g., for CPU and GPU); 

\item The block of the generative design of composite models, which implements the creation of models with specified properties by evolutionary algorithms. The properties of the models are determined by the target function passed to the optimizer. \ans{If there is more than one target function specified (as an example, the training time and modelling error can be used together as objectives for AutoML), then the multi-criteria formulation of the optimization problem is implemented, where the result of the model design is a Pareto front containing various compromising solutions}. The genotype is represented in graph form, and the crossing and mutation operators are implemented accordingly. 

\item The pipeline execution block on a given computational infrastructure. \ed{It allows individual pipeline execution on the given computational nodes.}  
\end{itemize}

This architecture is implemented in the core of the open-source FEDOT framework. Different aspects of its implementation are already detailed in a series of papers: 
 \cite{nikitin2020structural} describes the main schemes and the implementation of the evolutionary operators, \cite{polonskaia2021multi} is devoted to the multi-objective modification of this approach, and \cite{nikitin2021automated} provides an extended description of the various aspects of the evolutionary design for composite modelling pipelines. The tuning strategy of the pipeline hyperparameters is based on Bayesian optimization.
 
Custom models can be put inside this node. Search space for hyperparameters and initial approximations for the models should be specified manually if necessary. It makes it possible to involve the infrastructure-specific implementation of the model \ed{in} AutoML.

 The example below demonstrates the AutoML workflow from input data processing to obtaining prediction.

 \begin{lstlisting}[language=Python,basicstyle=\ttfamily,breaklines=true]
 api = Fedot(problem='classification', seed=42, timeout=30, preset='gpu')
 api.fit(features=x_train, target=y_train)
 predictions = api.predict(features=x_test)
 \end{lstlisting}
 
\begin{figure}[h!] 
    \centering
    \includegraphics[width=8cm]{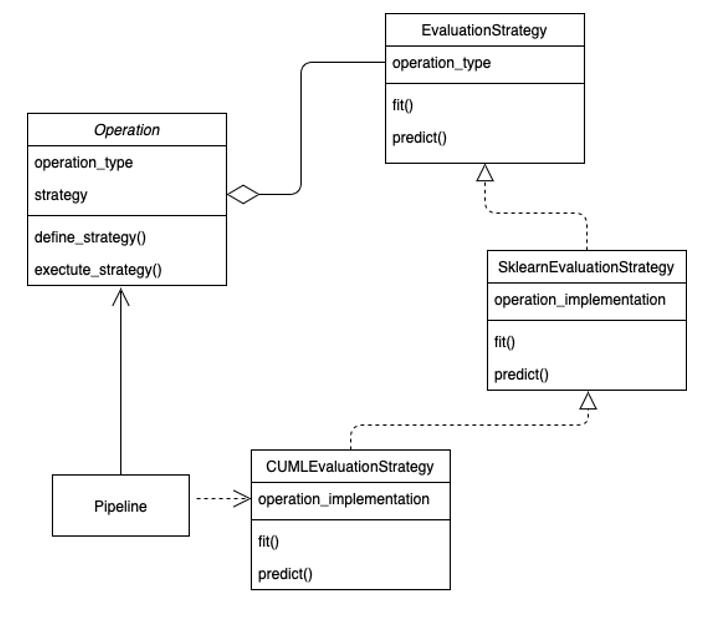}
    \caption{The class diagram \ans{for the implementation of a modelling pipeline that consists of several operations. The \textit{Operation} class represents a high-level modelling strategy that is used inside the operation. \textit{EvaluationStrategy} is a base class for the algorithmic implementation of this strategy. \textit{SklearnEvaluationStrategy} represents the implementation obtained from the scikit-learn library and \textit{CumlEvaluationStrategy} represents the implementation from the CUML library.}}
    \label{fig_class}
\end{figure}

Figure~\ref{fig_class} provides the UML class diagram for the implementation of various evaluation strategies that allow combining CPU- and GPU-based nodes in a single modelling pipeline. \ed{A high-level modelling method (e.g. Support Vector Classification) can be implemented using different algorithms: a CPU-optimised implementation of SVC can be obtained from the scikit-learn library \cite{scikit-learn}. In contrast, a GPU-optimised implementation is available in the CUML library. The proposed architecture makes it possible to hide these details inside the specific modelling pipeline and use the same optimisation logic for different implementations of the algorithms.}
 
\ans{The optimizer operates on individual models as a black box with input, output and fit/predict methods. The following building blocks can be used for pipelines: models (Bernoulli Naive Bayes classifier, logistic regression, multilayer perceptron, random forest, gradient boosting, k-nearest classifier, QDA, LDA, decision tree) and data transformation operations (scaling, normalization, polynomial features transformation, principal component analysis, independent component analysis, isolation forest, resampling).}

\section{\ans{Experimental Studies}}
\label{sec_exp}

We \ed{conducted} a series of experiments to confirm the correctness and effectiveness of the proposed approach. It can be divided into experiments with local and remote infrastructure. As benchmarks, various classification datasets from the OpenML base \cite{bischl2017openml} and \ans{synthetic datasets were used (the full list is presented in Table~\ref{tab_datas})}. A description of the computational infrastructure is provided for each experiment.

\begin{table}[ht!!]
\caption{\ans{The properties of OpenML datasets that were used during the experiments. The random forest model is used as a baseline for the training time estimation.}}
\label{tab_datas}
\begin{tabular}{|c|c|c|c|c|c|} 
\hline
\begin{tabular}[c]{@{}c@{}}\textbf{Dataset }\\\textbf{name}\end{tabular} & \begin{tabular}[c]{@{}c@{}}\textbf{Rows\textbf{,}}\\\textbf{\textbf{10\^3}}\end{tabular} & \textbf{Feat.} & \begin{tabular}[c]{@{}c@{}}\textbf{Total }\\\textbf{elem., }\\\textbf{10\^3}\end{tabular} & \begin{tabular}[c]{@{}c@{}}\textbf{Base.}\\\textbf{ train.}\\\textbf{time, }\\\textbf{ sec}\end{tabular} & \begin{tabular}[c]{@{}c@{}}\textbf{Num. }\\\textbf{ of }\\\textbf{ clas}\\\textbf{ses}\end{tabular}  \\ 
\hline
adult                                                                    & 49                                                                                       & 14             & 684                                                                                       & 12.5                                                                                                     & 2                                                                                                    \\ 
\hline
\begin{tabular}[c]{@{}c@{}}amazon\_\\employee\_\\access\end{tabular}     & 33                                                                                       & 9              & 295                                                                                       & 1.5                                                                                                      & 2                                                                                                    \\ 
\hline
australian                                                               & 0.69                                                                                     & 15             & 10                                                                                        & 0.2                                                                                                      & 2                                                                                                    \\ 
\hline
\begin{tabular}[c]{@{}c@{}}bank-\\marketing\end{tabular}                 & 45                                                                                       & 17             & 769                                                                                       & 6.9                                                                                                      & 2                                                                                                    \\ 
\hline
\begin{tabular}[c]{@{}c@{}}blood-\\transfusion\end{tabular}              & 0.75                                                                                     & 5              & 4                                                                                         & 0.1                                                                                                      & 2                                                                                                    \\ 
\hline
car                                                                      & 1,8                                                                                      & 7              & 12                                                                                        & 0.5                                                                                                      & 4                                                                                                    \\ 
\hline
cnae-9                                                                   & 1,1                                                                                      & 857            & 926                                                                                       & 14.7                                                                                                     & 9                                                                                                    \\ 
\hline
\begin{tabular}[c]{@{}c@{}}jungle\_chess\\\_2pcs\end{tabular}            & 45                                                                                       & 7              & 314                                                                                       & 48.0                                                                                                     & 3                                                                                                    \\ 
\hline
numerai28                                                                & 96                                                                                       & 22             & 2119                                                                                      & 8.4                                                                                                      & 2                                                                                                    \\ 
\hline
phoneme                                                                  & 54                                                                                       & 6              & 32                                                                                        & 0.12                                                                                                     & 2                                                                                                    \\ 
\hline
sylvine                                                                  & 51                                                                                       & 21             & 108                                                                                       & 0.5                                                                                                      & 2                                                                                                    \\ 
\hline
volkert                                                                  & 58                                                                                       & 181            & 10554                                                                                     & 128.4                                                                                                    & 10                                                                                                   \\ 
\hline
\begin{tabular}[c]{@{}c@{}}synthetic\_\\blobs\end{tabular}               & 100                                                                                      & 10             & 1000                                                                                      & 6.2                                                                                                      & 2                                                                                                    \\ 
\hline
\begin{tabular}[c]{@{}c@{}}synthetic\_\\moons\end{tabular}               & 1                                                                                        & 2              & 2                                                                                         & 0.12                                                                                                     & 2                                                                                                    \\
\hline
\end{tabular}
\end{table}

\ans{The following methodology was used for experimental studies: each experiment started with dividing samples into two groups: ‘learning’ and ‘validation’ samples in the ratio 70\% to 30\% to avoid data leaks. Then, the learning sample was transferred to the evolutionary optimizer. During the optimisation, the 5-fold cross-validation procedure was applied to estimate the values of the fitness function. 

The experiment is repeated three times for each dataset to take the stochasticity of the optimizer into account. The quality metrics are averaged over these iterations.}

\subsection{Local infrastructure}

For experiments with the local infrastructure, we configured a server based on Xeon Cascadelake (2900MHz) with 12 cores and 24Gb memory.

As our approach claims to increase the number of evaluated pipelines during fitting due to caching, it will be correct to compare this metric with and without the caching option. For that reason, we created the benchmark considering different computational setups for AutoML. It utilizes a dataset for classification present using the FEDOT framework as a test bench.

\begin{figure}[h!] 
    \centering
    \includegraphics[width=8cm]{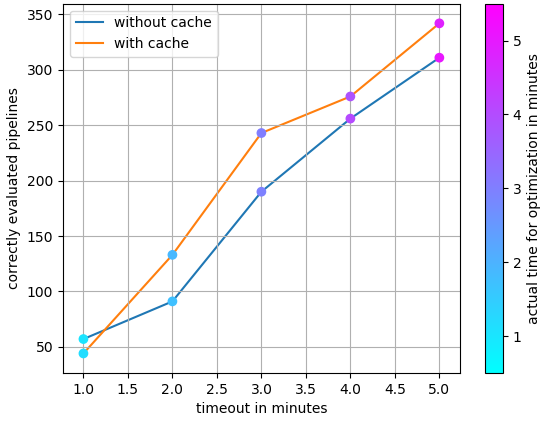}
    \caption{The dependence between the number of pipelines and the usage of cache (averaged for ten runs). Single-processing is used.}
    \label{fig_cache_exp_single}
\end{figure}

In Figure~\ref{fig_cache_exp_single} the comparison of cache-based and cache-free configurations is provided. For the first one, both the pipelines cache and data preprocessing cache are activated. The number of parallel jobs used during optimization is one. 

\ans{During evolutionary optimization, a lot of candidate solutions (pipelines) are evaluated. We repeated the experiment for different timeouts that limit the execution time for the entire AutoML run since they affect the number of evaluated pipelines. Also, an additional time limit is applied to the entire pipeline (to process the fit time anomalies for large pipelines). It is specified as 1/4 of the total timeout.}

Because of the stochastic nature of the optimization-based experiments, each run was repeated three times, and the obtained metrics were averaged.

\begin{figure}[h!] 
    \centering
    \includegraphics[width=8cm]{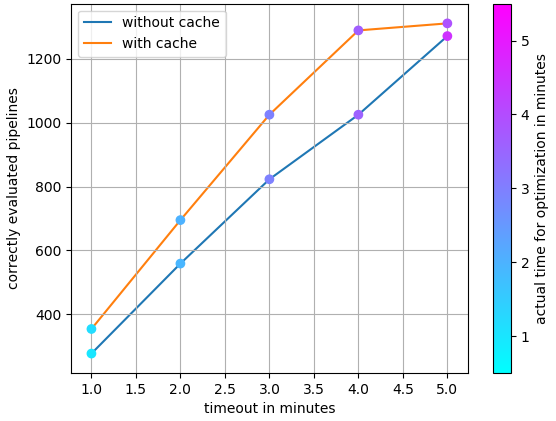}
    \caption{The dependence between the number of pipelines and the usage of cache (averaged for ten runs). 8 parallel jobs are used.}
    \label{fig_cache_exp_multi}
\end{figure}

The results presented in Figure~\ref{fig_cache_exp_multi} are obtained with the \textit{n\_jobs} \ans{hyperparameter} value \ed{equal} to \ans{12}.

Table~\ref{tab_cache} summarises the averaged metrics of the experiments with single-process and multi-process caching. The average performance was increased by 14 \%, which empirically confirms the effectiveness of the proposed approach.


\begin{table*}[]
\caption{\ans{The results of experiments with caching of pipeline nodes and 
 data preprocessing operations. The first column indicates whether the cache has been used, and the second column represents the number of parallel processes. The next three columns represent different metric values (the number of evaluated pipelines, ROC AUC for validation sample, and ROC AUC for cross-validation of training sample).}}
\renewcommand{\arraystretch}{0,46} 
\begin{tabular}{|c|cc|c|c|c|}
\hline
\multirow{2}{*}{\textbf{Dataset}}           & \multicolumn{2}{c|}{\textbf{Configuration}}                              & \multirow{2}{*}{\textbf{Pipelines count}} & \multirow{2}{*}{\textbf{ROC-AUC final}} & \multirow{2}{*}{\textbf{ROC-AUC cross-validation}} \\ \cline{2-3}
              & \multicolumn{1}{c|}{\textbf{Cache}}       & \textbf{Number of processes} &             &           &                      \\ \hline
\multirow{8}{*}{adult}                      & \multicolumn{1}{c|}{\multirow{2}{*}{on}}  & \multirow{4}{*}{1}           & \multirow{2}{*}{\textbf{27}}              & \multirow{2}{*}{0,92}                   & \multirow{2}{*}{0,9117}                            \\
              & \multicolumn{1}{c|}{}                     &                              &             &           &                      \\ \cline{2-2} \cline{4-6} 
              & \multicolumn{1}{c|}{\multirow{2}{*}{off}} &                              & \multirow{2}{*}{23}                       & \multirow{2}{*}{\textbf{0,9213}}        & \multirow{2}{*}{\textbf{0,913}}                    \\
              & \multicolumn{1}{c|}{}                     &                              &             &           &                      \\ \cline{2-6} 
              & \multicolumn{1}{c|}{\multirow{2}{*}{on}}  & \multirow{4}{*}{8}           & \multirow{2}{*}{\textbf{190}}             & \multirow{2}{*}{0,921}                  & \multirow{2}{*}{0,9131}                            \\
              & \multicolumn{1}{c|}{}                     &                              &             &           &                      \\ \cline{2-2} \cline{4-6} 
              & \multicolumn{1}{c|}{\multirow{2}{*}{off}} &                              & \multirow{2}{*}{170}                      & \multirow{2}{*}{\textbf{0,922}}         & \multirow{2}{*}{\textbf{0,9137}}                   \\
              & \multicolumn{1}{c|}{}                     &                              &             &           &                      \\ \hline
\multirow{8}{*}{amazon\_employee\_access}                                 & \multicolumn{1}{c|}{\multirow{2}{*}{on}}  & \multirow{4}{*}{1}           & \multirow{2}{*}{\textbf{85}}              & \multirow{2}{*}{0,8447}                 & \multirow{2}{*}{0,8346}                            \\
              & \multicolumn{1}{c|}{}                     &                              &             &           &                      \\ \cline{2-2} \cline{4-6} 
              & \multicolumn{1}{c|}{\multirow{2}{*}{off}} &                              & \multirow{2}{*}{78}                       & \multirow{2}{*}{\textbf{0,8497}}        & \multirow{2}{*}{\textbf{0,8376}}                   \\
              & \multicolumn{1}{c|}{}                     &                              &             &           &                      \\ \cline{2-6} 
              & \multicolumn{1}{c|}{\multirow{2}{*}{on}}  & \multirow{4}{*}{8}           & \multirow{2}{*}{\textbf{416}}             & \multirow{2}{*}{\textbf{0,8507}}        & \multirow{2}{*}{0,8356}                            \\
              & \multicolumn{1}{c|}{}                     &                              &             &           &                      \\ \cline{2-2} \cline{4-6} 
              & \multicolumn{1}{c|}{\multirow{2}{*}{off}} &                              & \multirow{2}{*}{369}                      & \multirow{2}{*}{0,849}                  & \multirow{2}{*}{\textbf{0,8398}}                   \\
              & \multicolumn{1}{c|}{}                     &                              &             &           &                      \\ \hline
\multirow{8}{*}{australian}                 & \multicolumn{1}{c|}{\multirow{2}{*}{on}}  & \multirow{4}{*}{1}           & \multirow{2}{*}{\textbf{879}}             & \multirow{2}{*}{\textbf{0,9313}}        & \multirow{2}{*}{\textbf{0,9432}}                   \\
              & \multicolumn{1}{c|}{}                     &                              &             &           &                      \\ \cline{2-2} \cline{4-6} 
              & \multicolumn{1}{c|}{\multirow{2}{*}{off}} &                              & \multirow{2}{*}{838}                      & \multirow{2}{*}{0,9283}                 & \multirow{2}{*}{0,9401}                            \\
              & \multicolumn{1}{c|}{}                     &                              &             &           &                      \\ \cline{2-6} 
              & \multicolumn{1}{c|}{\multirow{2}{*}{on}}  & \multirow{4}{*}{8}           & \multirow{2}{*}{\textbf{6354}}            & \multirow{2}{*}{0,928}                  & \multirow{2}{*}{0,9411}                            \\
              & \multicolumn{1}{c|}{}                     &                              &             &           &                      \\ \cline{2-2} \cline{4-6} 
              & \multicolumn{1}{c|}{\multirow{2}{*}{off}} &                              & \multirow{2}{*}{6199}                     & \multirow{2}{*}{\textbf{0,934}}         & \multirow{2}{*}{\textbf{0,9442}}                   \\
              & \multicolumn{1}{c|}{}                     &                              &             &           &                      \\ \hline
\multirow{8}{*}{bank-marketing}             & \multicolumn{1}{c|}{\multirow{2}{*}{on}}  & \multirow{4}{*}{1}           & \multirow{2}{*}{\textbf{38}}              & \multirow{2}{*}{0,93}                   & \multirow{2}{*}{\textbf{0,931}}                    \\
              & \multicolumn{1}{c|}{}                     &                              &             &           &                      \\ \cline{2-2} \cline{4-6} 
              & \multicolumn{1}{c|}{\multirow{2}{*}{off}} &                              & \multirow{2}{*}{30}                       & \multirow{2}{*}{\textbf{0,9313}}        & \multirow{2}{*}{0,93}                              \\
              & \multicolumn{1}{c|}{}                     &                              &             &           &                      \\ \cline{2-6} 
              & \multicolumn{1}{c|}{\multirow{2}{*}{on}}  & \multirow{4}{*}{8}           & \multirow{2}{*}{205}                      & \multirow{2}{*}{0,931}                  & \multirow{2}{*}{\textbf{0,932}}                    \\
              & \multicolumn{1}{c|}{}                     &                              &             &           &                      \\ \cline{2-2} \cline{4-6} 
              & \multicolumn{1}{c|}{\multirow{2}{*}{off}} &                              & \multirow{2}{*}{\textbf{211}}             & \multirow{2}{*}{\textbf{0,932}}         & \multirow{2}{*}{0,931}                             \\
              & \multicolumn{1}{c|}{}                     &                              &             &           &                      \\ \hline
\multirow{8}{*}{\begin{tabular}[c]{@{}c@{}}blood-transfusion\\ -service-center\end{tabular}}            & \multicolumn{1}{c|}{\multirow{2}{*}{on}}  & \multirow{4}{*}{1}           & \multirow{2}{*}{\textbf{2175}}            & \multirow{2}{*}{\textbf{0,748}}         & \multirow{2}{*}{0,75}                              \\
              & \multicolumn{1}{c|}{}                     &                              &             &           &                      \\ \cline{2-2} \cline{4-6} 
              & \multicolumn{1}{c|}{\multirow{2}{*}{off}} &                              & \multirow{2}{*}{2064}                     & \multirow{2}{*}{0,7383}                 & \multirow{2}{*}{\textbf{0,759}}                    \\
              & \multicolumn{1}{c|}{}                     &                              &             &           &                      \\ \cline{2-6} 
              & \multicolumn{1}{c|}{\multirow{2}{*}{on}}  & \multirow{4}{*}{8}           & \multirow{2}{*}{\textbf{13943}}           & \multirow{2}{*}{0,745}                  & \multirow{2}{*}{0,761}                             \\
              & \multicolumn{1}{c|}{}                     &                              &             &           &                      \\ \cline{2-2} \cline{4-6} 
              & \multicolumn{1}{c|}{\multirow{2}{*}{off}} &                              & \multirow{2}{*}{13834}                    & \multirow{2}{*}{\textbf{0,749}}         & \multirow{2}{*}{\textbf{0,7659}}                   \\
              & \multicolumn{1}{c|}{}                     &                              &             &           &                      \\ \hline
\multirow{8}{*}{car}                        & \multicolumn{1}{c|}{\multirow{2}{*}{on}}  & \multirow{4}{*}{1}           & \multirow{2}{*}{\textbf{812}}             & \multirow{2}{*}{0,921}                  & \multirow{2}{*}{\textbf{0,933}}                    \\
              & \multicolumn{1}{c|}{}                     &                              &             &           &                      \\ \cline{2-2} \cline{4-6} 
              & \multicolumn{1}{c|}{\multirow{2}{*}{off}} &                              & \multirow{2}{*}{728}                      & \multirow{2}{*}{\textbf{0,9233}}        & \multirow{2}{*}{0,9319}                            \\
              & \multicolumn{1}{c|}{}                     &                              &             &           &                      \\ \cline{2-6} 
              & \multicolumn{1}{c|}{\multirow{2}{*}{on}}  & \multirow{4}{*}{8}           & \multirow{2}{*}{\textbf{4856}}            & \multirow{2}{*}{\textbf{0,922}}         & \multirow{2}{*}{\textbf{0,935}}                    \\
              & \multicolumn{1}{c|}{}                     &                              &             &           &                      \\ \cline{2-2} \cline{4-6} 
              & \multicolumn{1}{c|}{\multirow{2}{*}{off}} &                              & \multirow{2}{*}{4608}                     & \multirow{2}{*}{0,92}                   & \multirow{2}{*}{0,934}                             \\
              & \multicolumn{1}{c|}{}                     &                              &             &           &                      \\ \hline
\multirow{8}{*}{cnae-9}                     & \multicolumn{1}{c|}{\multirow{2}{*}{on}}  & \multirow{4}{*}{1}           & \multirow{2}{*}{\textbf{214}}             & \multirow{2}{*}{\textbf{0,995}}         & \multirow{2}{*}{\textbf{0,9939}}                   \\
              & \multicolumn{1}{c|}{}                     &                              &             &           &                      \\ \cline{2-2} \cline{4-6} 
              & \multicolumn{1}{c|}{\multirow{2}{*}{off}} &                              & \multirow{2}{*}{195}                      & \multirow{2}{*}{\textbf{0,995}}         & \multirow{2}{*}{\textbf{0,9939}}                   \\
              & \multicolumn{1}{c|}{}                     &                              &             &           &                      \\ \cline{2-6} 
              & \multicolumn{1}{c|}{\multirow{2}{*}{on}}  & \multirow{4}{*}{8}           & \multirow{2}{*}{1100}                     & \multirow{2}{*}{\textbf{0,995}}         & \multirow{2}{*}{0,9942}                            \\
              & \multicolumn{1}{c|}{}                     &                              &             &           &                      \\ \cline{2-2} \cline{4-6} 
              & \multicolumn{1}{c|}{\multirow{2}{*}{off}} &                              & \multirow{2}{*}{\textbf{1161}}            & \multirow{2}{*}{\textbf{0,995}}         & \multirow{2}{*}{\textbf{0,9953}}                   \\
              & \multicolumn{1}{c|}{}                     &                              &             &           &                      \\ \hline
\multirow{8}{*}{\begin{tabular}[c]{@{}c@{}}jungle\_chess\_2pcs\_raw\\ \_endgame\_complete\end{tabular}} & \multicolumn{1}{c|}{\multirow{2}{*}{on}}  & \multirow{4}{*}{1}           & \multirow{2}{*}{30}                       & \multirow{2}{*}{\textbf{0,9671}}        & \multirow{2}{*}{\textbf{0,9637}}                   \\
              & \multicolumn{1}{c|}{}                     &                              &             &           &                      \\ \cline{2-2} \cline{4-6} 
              & \multicolumn{1}{c|}{\multirow{2}{*}{off}} &                              & \multirow{2}{*}{\textbf{44}}              & \multirow{2}{*}{0,9667}                 & \multirow{2}{*}{0,9627}                            \\
              & \multicolumn{1}{c|}{}                     &                              &             &           &                      \\ \cline{2-6} 
              & \multicolumn{1}{c|}{\multirow{2}{*}{on}}  & \multirow{4}{*}{8}           & \multirow{2}{*}{89}                       & \multirow{2}{*}{0,969}                  & \multirow{2}{*}{0,9631}                            \\
              & \multicolumn{1}{c|}{}                     &                              &             &           &                      \\ \cline{2-2} \cline{4-6} 
              & \multicolumn{1}{c|}{\multirow{2}{*}{off}} &                              & \multirow{2}{*}{\textbf{99}}              & \multirow{2}{*}{\textbf{0,9713}}        & \multirow{2}{*}{\textbf{0,9649}}                   \\
              & \multicolumn{1}{c|}{}                     &                              &             &           &                      \\ \hline
\multirow{8}{*}{numerai28}                  & \multicolumn{1}{c|}{\multirow{2}{*}{on}}  & \multirow{4}{*}{1}           & \multirow{2}{*}{3}                        & \multirow{2}{*}{0.508}                  & \multirow{2}{*}{0,51}                              \\
              & \multicolumn{1}{c|}{}                     &                              &             &           &                      \\ \cline{2-2} \cline{4-6} 
              & \multicolumn{1}{c|}{\multirow{2}{*}{off}} &                              & \multirow{2}{*}{\textbf{6}}               & \multirow{2}{*}{\textbf{0,511}}         & \multirow{2}{*}{\textbf{0,5182}}                   \\
              & \multicolumn{1}{c|}{}                     &                              &             &           &                      \\ \cline{2-6} 
              & \multicolumn{1}{c|}{\multirow{2}{*}{on}}  & \multirow{4}{*}{8}           & \multirow{2}{*}{20}                       & \multirow{2}{*}{0,527}                  & \multirow{2}{*}{\textbf{0,528}}                    \\
              & \multicolumn{1}{c|}{}                     &                              &             &           &                      \\ \cline{2-2} \cline{4-6} 
              & \multicolumn{1}{c|}{\multirow{2}{*}{off}} &                              & \multirow{2}{*}{\textbf{23}}              & \multirow{2}{*}{\textbf{0,5273}}        & \multirow{2}{*}{\textbf{0,528}}                    \\
              & \multicolumn{1}{c|}{}                     &                              &             &           &                      \\ \hline
\multirow{8}{*}{phoneme}                    & \multicolumn{1}{c|}{\multirow{2}{*}{on}}  & \multirow{4}{*}{1}           & \multirow{2}{*}{\textbf{354}}             & \multirow{2}{*}{\textbf{0,9599}}        & \multirow{2}{*}{0,951}                             \\
              & \multicolumn{1}{c|}{}                     &                              &             &           &                      \\ \cline{2-2} \cline{4-6} 
              & \multicolumn{1}{c|}{\multirow{2}{*}{off}} &                              & \multirow{2}{*}{325}                      & \multirow{2}{*}{0,9597}                 & \multirow{2}{*}{\textbf{0,9515}}                   \\
              & \multicolumn{1}{c|}{}                     &                              &             &           &                      \\ \cline{2-6} 
              & \multicolumn{1}{c|}{\multirow{2}{*}{on}}  & \multirow{4}{*}{8}           & \multirow{2}{*}{\textbf{1954}}            & \multirow{2}{*}{\textbf{0,9631}}        & \multirow{2}{*}{\textbf{0,955}}                    \\
              & \multicolumn{1}{c|}{}                     &                              &             &           &                      \\ \cline{2-2} \cline{4-6} 
              & \multicolumn{1}{c|}{\multirow{2}{*}{off}} &                              & \multirow{2}{*}{1885}                     & \multirow{2}{*}{0,963}                  & \multirow{2}{*}{0,9547}                            \\
              & \multicolumn{1}{c|}{}                     &                              &             &           &                      \\ \hline
\multirow{8}{*}{sylvine}                    & \multicolumn{1}{c|}{\multirow{2}{*}{on}}  & \multirow{4}{*}{1}           & \multirow{2}{*}{\textbf{221}}             & \multirow{2}{*}{0,9852}                 & \multirow{2}{*}{\textbf{0,9809}}                   \\
              & \multicolumn{1}{c|}{}                     &                              &             &           &                      \\ \cline{2-2} \cline{4-6} 
              & \multicolumn{1}{c|}{\multirow{2}{*}{off}} &                              & \multirow{2}{*}{206}                      & \multirow{2}{*}{\textbf{0,9853}}        & \multirow{2}{*}{0,9806}                            \\
              & \multicolumn{1}{c|}{}                     &                              &             &           &                      \\ \cline{2-6} 
              & \multicolumn{1}{c|}{\multirow{2}{*}{on}}  & \multirow{4}{*}{8}           & \multirow{2}{*}{\textbf{932}}             & \multirow{2}{*}{\textbf{0,9878}}        & \multirow{2}{*}{0,981}                             \\
              & \multicolumn{1}{c|}{}                     &                              &             &           &                      \\ \cline{2-2} \cline{4-6} 
              & \multicolumn{1}{c|}{\multirow{2}{*}{off}} &                              & \multirow{2}{*}{827}                      & \multirow{2}{*}{0,9877}                 & \multirow{2}{*}{\textbf{0,9829}}                   \\
              & \multicolumn{1}{c|}{}                     &                              &             &           &                      \\ \hline
\multirow{8}{*}{volkert}                    & \multicolumn{1}{c|}{\multirow{2}{*}{on}}  & \multirow{4}{*}{1}           & \multirow{2}{*}{4}                        & \multirow{2}{*}{0,932}                  & \multirow{2}{*}{0,9298}                            \\
              & \multicolumn{1}{c|}{}                     &                              &             &           &                      \\ \cline{2-2} \cline{4-6} 
              & \multicolumn{1}{c|}{\multirow{2}{*}{off}} &                              & \multirow{2}{*}{\textbf{6}}               & \multirow{2}{*}{\textbf{0,9393}}        & \multirow{2}{*}{\textbf{0,9344}}                   \\
              & \multicolumn{1}{c|}{}                     &                              &             &           &                      \\ \cline{2-6} 
              & \multicolumn{1}{c|}{\multirow{2}{*}{on}}  & \multirow{4}{*}{8}           & \multirow{2}{*}{20}                       & \multirow{2}{*}{0,9313}                 & \multirow{2}{*}{\textbf{0,934}}                    \\
              & \multicolumn{1}{c|}{}                     &                              &             &           &                      \\ \cline{2-2} \cline{4-6} 
              & \multicolumn{1}{c|}{\multirow{2}{*}{off}} &                              & \multirow{2}{*}{\textbf{21}}              & \multirow{2}{*}{\textbf{0,9317}}        & \multirow{2}{*}{0,9273}                            \\
              & \multicolumn{1}{c|}{}                     &                              &             &           &                      \\ \hline

\multirow{8}{*}{synthetic\_blobs}                      & \multicolumn{1}{c|}{\multirow{2}{*}{on}}  & \multirow{4}{*}{1}           & \multirow{2}{*}{\textbf{31}}            & \multirow{2}{*}{\textbf{1}}             & \multirow{2}{*}{\textbf{1}}                        \\
              & \multicolumn{1}{c|}{}                     &                              &             &           &                      \\ \cline{2-2} \cline{4-6} 
              & \multicolumn{1}{c|}{\multirow{2}{*}{off}} &                              & \multirow{2}{*}{27}                     & \multirow{2}{*}{\textbf{1}}             & \multirow{2}{*}{\textbf{1}}                        \\
              & \multicolumn{1}{c|}{}                     &                              &             &           &                      \\ \cline{2-6} 
              & \multicolumn{1}{c|}{\multirow{2}{*}{on}}  & \multirow{4}{*}{8}           & \multirow{2}{*}{\textbf{235}}            & \multirow{2}{*}{\textbf{1}}             & \multirow{2}{*}{\textbf{1}}                        \\
              & \multicolumn{1}{c|}{}                     &                              &             &           &                      \\ \cline{2-2} \cline{4-6} 
              & \multicolumn{1}{c|}{\multirow{2}{*}{off}} &                              & \multirow{2}{*}{224}                     & \multirow{2}{*}{\textbf{1}}             & \multirow{2}{*}{\textbf{1}}                        \\
              & \multicolumn{1}{c|}{}                     &                              &             &           &                      \\ \hline
\multirow{8}{*}{synthetic\_moons}                      & \multicolumn{1}{c|}{\multirow{2}{*}{on}}  & \multirow{4}{*}{1}           & \multirow{2}{*}{1124}                     & \multirow{2}{*}{\textbf{1}}             & \multirow{2}{*}{\textbf{1}}                        \\
              & \multicolumn{1}{c|}{}                     &                              &             &           &                      \\ \cline{2-2} \cline{4-6} 
              & \multicolumn{1}{c|}{\multirow{2}{*}{off}} &                              & \multirow{2}{*}{\textbf{1026}}            & \multirow{2}{*}{\textbf{1}}             & \multirow{2}{*}{\textbf{1}}                        \\
              & \multicolumn{1}{c|}{}                     &                              &             &           &                      \\ \cline{2-6} 
              & \multicolumn{1}{c|}{\multirow{2}{*}{on}}  & \multirow{4}{*}{8}           & \multirow{2}{*}{12356}                    & \multirow{2}{*}{\textbf{1}}             & \multirow{2}{*}{\textbf{1}}                        \\
              & \multicolumn{1}{c|}{}                     &                              &             &           &                      \\ \cline{2-2} \cline{4-6} 
              & \multicolumn{1}{c|}{\multirow{2}{*}{off}} &                              & \multirow{2}{*}{\textbf{12227}}           & \multirow{2}{*}{\textbf{1}}             & \multirow{2}{*}{\textbf{1}}                        \\
              & \multicolumn{1}{c|}{}                     &                              &             &           &                      \\ \hline

\end{tabular}
\label{tab_cache}
\end{table*}

The next stage of the experiment is devoted to the analysis of the evolutionary algorithm's performance in multiprocessing mode. We compared algorithm performance with the number of processes equal to 1 and 8 with a timeout \ed{set} to 10 minutes. \ans{The optimization of the pipeline structure was repeated three times with no seed and with five cross validation folds to take stochasticity into account.} 

\begin{figure}[h!] 
    \centering
    \includegraphics[width=8cm]{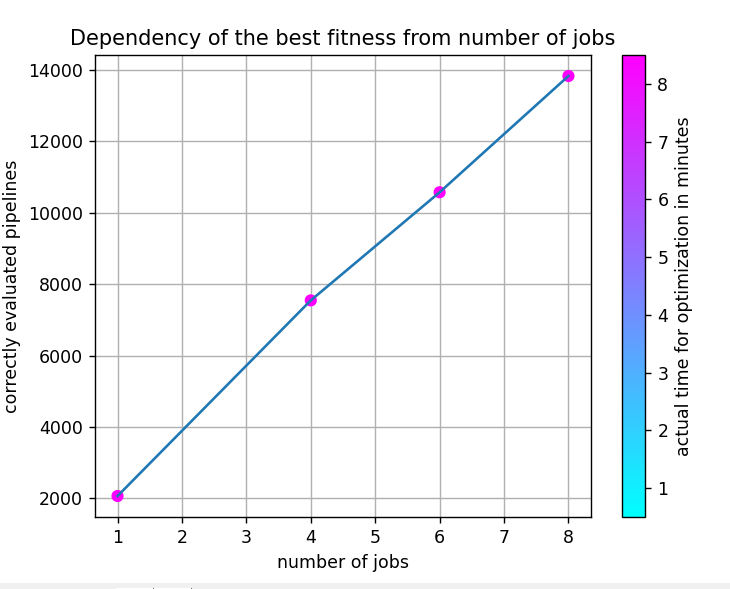}
    \caption{The detailed analysis of dependency of number of pipelines from the number of jobs (dataset blood-transfusion-service-center)}
    \label{fig_exp_multi}
\end{figure}

\begin{figure}[h!] 
    \centering
    \includegraphics[width=8cm]{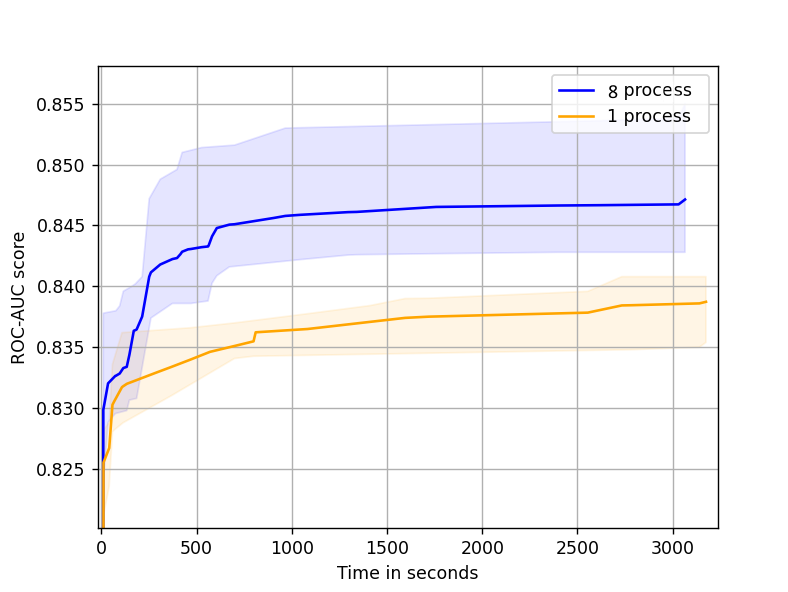}
    \caption{\ed{The dependency of the best fitness values on the optimization time}. The intervals represent the stochasticity of the optimization runs.}
    \label{fig_exp_multi_time}
\end{figure}

The dependency of correctly evaluated pipelines on a specified number of jobs for a single dataset is presented in Figure~\ref{fig_exp_multi}. It can be seen that near-linear improvement in parallel speedup is achieved. Figure~\ref{fig_exp_multi_time} \ed{demonstrates the dependency of the best fitness calculated using cross-validation on the timestamp from the configuration.} Launches with 8 processes find a better solution faster than launches with one process.

Table~\ref{tab_multi} summarises the averaged results of experiments \ed{in} single-process and multiprocessing modes. The fitness score \ed{calculated using} cross-validation increases linearly with the number of evaluated pipelines. It confirms the effectiveness of the local parallelization of evolutionary AutoML.

\begin{table*}[]
\caption{\ans{The results of experiments with parallelization of evolution. The first two rows for each dataset (1 and 8 jobs) represent the results obtained with a fit time limit for pipelines, while the "without limit" row contain the results obtained without limits with 8 jobs.}}
\begin{tabular}{|c|c|c|c|c|}
\hline
\textbf{Dataset}               & \textbf{Number of processes} & \textbf{\begin{tabular}[c]{@{}c@{}}Pipelines\\  count\end{tabular}} & \textbf{\begin{tabular}[c]{@{}c@{}}ROC-AUC\\  final\end{tabular}} & \textbf{\begin{tabular}[c]{@{}c@{}}ROC-AUC\\  cross-validation\end{tabular}} \\ \hline
\multirow{3}{*}{adult}         & 1 (with limit)     & 23      & 0,9213       & 0,913            \\ \cline{2-5} 
        & 8 (with limit)     & \textbf{170}                          & \textbf{0,922}                      & 0,9137           \\ \cline{2-5} 
        & 8 (without limit)  & \multicolumn{1}{c|}{126}              & \multicolumn{1}{c|}{0,9217}         & \textbf{0,9141}         \\ \hline
\multirow{3}{*}{amazon\_employee\_access}                    & 1 (with limit)     & 78      & 0,8497       & 0,8376           \\ \cline{2-5} 
        & 8 (with limit)     & \textbf{369}                          & \textbf{0,849}                      & \textbf{0,8398}         \\ \cline{2-5} 
        & 8 (without limit)  & 329            & \textbf{0,849}                      & 0,8387           \\ \hline
\multirow{3}{*}{australian}    & 1 (with limit)     & 838            & 0,9283       & 0,9401           \\ \cline{2-5} 
        & 8 (with limit)     & 6199           & \textbf{0,934}                      & 0,9442           \\ \cline{2-5} 
        & 8 (without limit)  & \textbf{10009}                        & 0,9307       & \textbf{0,9444}         \\ \hline
\multirow{3}{*}{bank-marketing}       & 1 (with limit)     & 30      & 0,9313       & 0,93             \\ \cline{2-5} 
        & 8 (with limit)     & \textbf{211}                          & \textbf{0,932}                      & \textbf{0,931}          \\ \cline{2-5} 
        & 8 (without limit)  & 184            & 0,9313       & 0,93             \\ \hline
\multirow{3}{*}{blood-transfusion-service-center}            & 1 (with limit)     & 2064           & 0,7383       & 0,759            \\ \cline{2-5} 
        & 8 (with limit)     & \textbf{13834}                        & \textbf{0,749}                      & \textbf{0,7659}         \\ \cline{2-5} 
        & 8 (without limit)  & 8329           & 0,716        & 0,7658           \\ \hline
\multirow{3}{*}{car}           & 1 (with limit)     & 728            & \textbf{0,9233}                     & 0,9319           \\ \cline{2-5} 
        & 8 (with limit)     & 4608           & 0,92         & 0,934            \\ \cline{2-5} 
        & 8 (without limit)  & \textbf{4612}                         & \textbf{0,925}                      & \textbf{0,9345}         \\ \hline
\multirow{3}{*}{cnae-9}        & 1 (with limit)     & 195            & 0,995        & 0,9939           \\ \cline{2-5} 
        & 8 (with limit)     & 1161           & 0,995        & \textbf{0,9953}         \\ \cline{2-5} 
        & 8 (without limit)  & \textbf{1168}                         & \textbf{0,995}                      & 0,9949           \\ \hline
\multirow{3}{*}{jungle\_chess\_2pcs\_raw\_endgame\_complete} & 1 (with limit)     & 44      & 0,9667       & 0,9627           \\ \cline{2-5} 
        & 8 (with limit)     & 99      & 0,9713       & 0,9649           \\ \cline{2-5} 
        & 8 (without limit)  & \textbf{144}                          & \textbf{0,9723}                     & \textbf{0,9661}         \\ \hline
\multirow{3}{*}{numerai28}     & 1 (with limit)     & 6       & 0,511        & 0,5182           \\ \cline{2-5} 
        & 8 (with limit)     & 23      & \textbf{0,5273}                     & \textbf{0,528}          \\ \cline{2-5} 
        & 8 (without limit)  & \textbf{22}    & 0,5253       & \textbf{0,5266}         \\ \hline
\multirow{3}{*}{phoneme}       & 1 (with limit)     & 325            & 0,9597       & 0,9515           \\ \cline{2-5} 
        & 8 (with limit)     & 1885           & \textbf{0,963}                      & \textbf{0,9547}         \\ \cline{2-5} 
        & 8 (without limit)  & \textbf{1917}                         & \textbf{0,963}                      & 0,9536           \\ \hline
\multirow{3}{*}{sylvine}       & 1 (with limit)     & 206            & 0,9853       & 0,9806           \\ \cline{2-5} 
        & 8 (with limit)     & \textbf{827}                          & \textbf{0,9877}                     & \textbf{0,9829}         \\ \cline{2-5} 
        & 8 (without limit)  & 816            & 0,986        & 0,9821           \\ \hline
\multirow{3}{*}{volkert}       & 1 (with limit)     & 6       & 0,9373       & 0,9329           \\ \cline{2-5} 
        & 8 (with limit)     & \textbf{21}    & \textbf{0,9393}                     & \textbf{0,9344}         \\ \cline{2-5} 
        & 8 (without limit)  & \textbf{21}    & 0,9317       & 0,9273           \\ \hline
\multirow{3}{*}{synthetic\_blobs}         & 1 (with limit)     & 8           & \textbf{1}                          & \textbf{1}       \\ \cline{2-5} 
        & 8 (with limit)     & \textbf{37}                        & \textbf{1}                          & \textbf{1}       \\ \cline{2-5} 
        & 8 (without limit)  & 30                        & \textbf{1}                          & \textbf{1}       \\ \hline
\multirow{3}{*}{synthetic\_moons}         & 1 (with limit)     & 1026           & \textbf{1}                          & \textbf{1}       \\ \cline{2-5} 
        & 8 (with limit)     & 12227          & \textbf{1}                          & \textbf{1}       \\ \cline{2-5} 
        & 8 (without limit)  & \textbf{12569}                        & \textbf{1}                          & \textbf{1}       \\ \hline

\end{tabular}
\label{tab_multi}
\end{table*}

\subsection{Heterogeneous infrastructure}

In the next series of experiments, we aim to estimate the efficiency of heterogeneous infrastructure for large datasets. Computation experiments were performed in a supercomputer environment configured based on two DGX-1 clusters. Each cluster contains eight Tesla V100 graphics cards and 128 GB of video memory. The number of graphics cores is 40960. 

The first experiment compares AutoML performance for the CPU-only and the hybrid infrastructures for various tasks. \ans{The aim of the experiment is to estimate the decreasing of fitting time after involvement of GPU-based nodes. To reduce the computational complexity of experiments, we decided not to use the full set of datasets from Table~\ref{tab_datas}. In this experiment, four synthetic binary classification datasets with 10 features and different number of rows (10000, 100000, 200000 and and 300000 rows).} Both single-model (SVC) and multi-model pipelines (consisting of SVC, Logistic Regression, and Random Forest) are considered to estimate the overhead for data flow transfer between models in the pipeline. The results are presented in Table~\ref{tab_gpu}.

\begin{table}[]
\caption{The training time of the pipelines on synthetic data under different conditions (single SVC classifier and composite pipeline with several models are considered). Averaged efficiency estimations are presented for homogeneous (one server with multi-core CPU) and heterogeneous (CPU and GPU) computing environments.}
\begin{tabular}{|c|cccc|cc|}
\hline
\multirow{3}{*}{{\rotatebox[origin=c]{90}{Rows, $10^3$}}} & \multicolumn{4}{c|}{Fitting time, sec}        & \multicolumn{2}{c|}{\begin{tabular}[c]{@{}c@{}}Improvement, \\ \%\end{tabular}}          \\ \cline{2-7} 
       & \multicolumn{2}{c|}{\begin{tabular}[c]{@{}c@{}}Single\\ model\end{tabular}}     & \multicolumn{2}{c|}{\begin{tabular}[c]{@{}c@{}}Comp.\\ pipeline\end{tabular}}  & \multicolumn{1}{c|}{\multirow{2}{*}{\begin{tabular}[c]{@{}c@{}}Single\\ model\end{tabular}}} & \multirow{2}{*}{\begin{tabular}[c]{@{}c@{}}Comp.\\ pipeline\end{tabular}} \\ \cline{2-5}
       & \multicolumn{1}{c|}{CPU}  & \multicolumn{1}{c|}{\begin{tabular}[c]{@{}c@{}}CPU+\\ GPU\end{tabular}} & \multicolumn{1}{c|}{CPU}  & \begin{tabular}[c]{@{}c@{}}CPU+\\ GPU\end{tabular} & \multicolumn{1}{c|}{}  &     \\ \hline
10     & \multicolumn{1}{c|}{0.3}  & \multicolumn{1}{c|}{2.2}   & \multicolumn{1}{c|}{0.7}  & 2.4   & \multicolumn{1}{c|}{-}      & -   \\ \hline
100    & \multicolumn{1}{c|}{11.4} & \multicolumn{1}{c|}{1.9}   & \multicolumn{1}{c|}{8.3}  & 1.6   & \multicolumn{1}{c|}{91}      & 88  \\ \hline
200    & \multicolumn{1}{c|}{39.3} & \multicolumn{1}{c|}{2.8}   & \multicolumn{1}{c|}{21}   & 3.2   & \multicolumn{1}{c|}{94}      & 85  \\ \hline
300    & \multicolumn{1}{c|}{76.0} & \multicolumn{1}{c|}{4.2}   & \multicolumn{1}{c|}{37.8} & 5.5   & \multicolumn{1}{c|}{95}      & 86  \\ \hline
\end{tabular}
\label{tab_gpu}
\end{table}

The results confirmed that the overhead could exceed the performance gain for a small amount of data. However, the proposed hybrid approach to pipeline evaluation is reasonably practical for large datasets.

\subsection{Remote infrastructure}
The next series of experiments uses a homogeneous cluster of 20 nodes under Kubernetes control. Each node has 40 CPU cores and 256 Gb RAM. We have trained populations of 50, 100 and 200 individuals. \ans{Each population has been trained four times, then the estimated time values were averaged. The aim of this experiment is to analyze the structure of computing time for remote evaluation and confirm that remote evaluation can be viable for large datasets regardless of existing overheads. To make the results of the experiment more compact, we focused on an analysis of a single synthetic binary classification dataset with 300000 rows and 10 features.}

\begin{figure*}[ht!]
     \centering
     \begin{subfigure}[b]{0.48\textwidth}
         \centering
         \includegraphics[width=\linewidth]{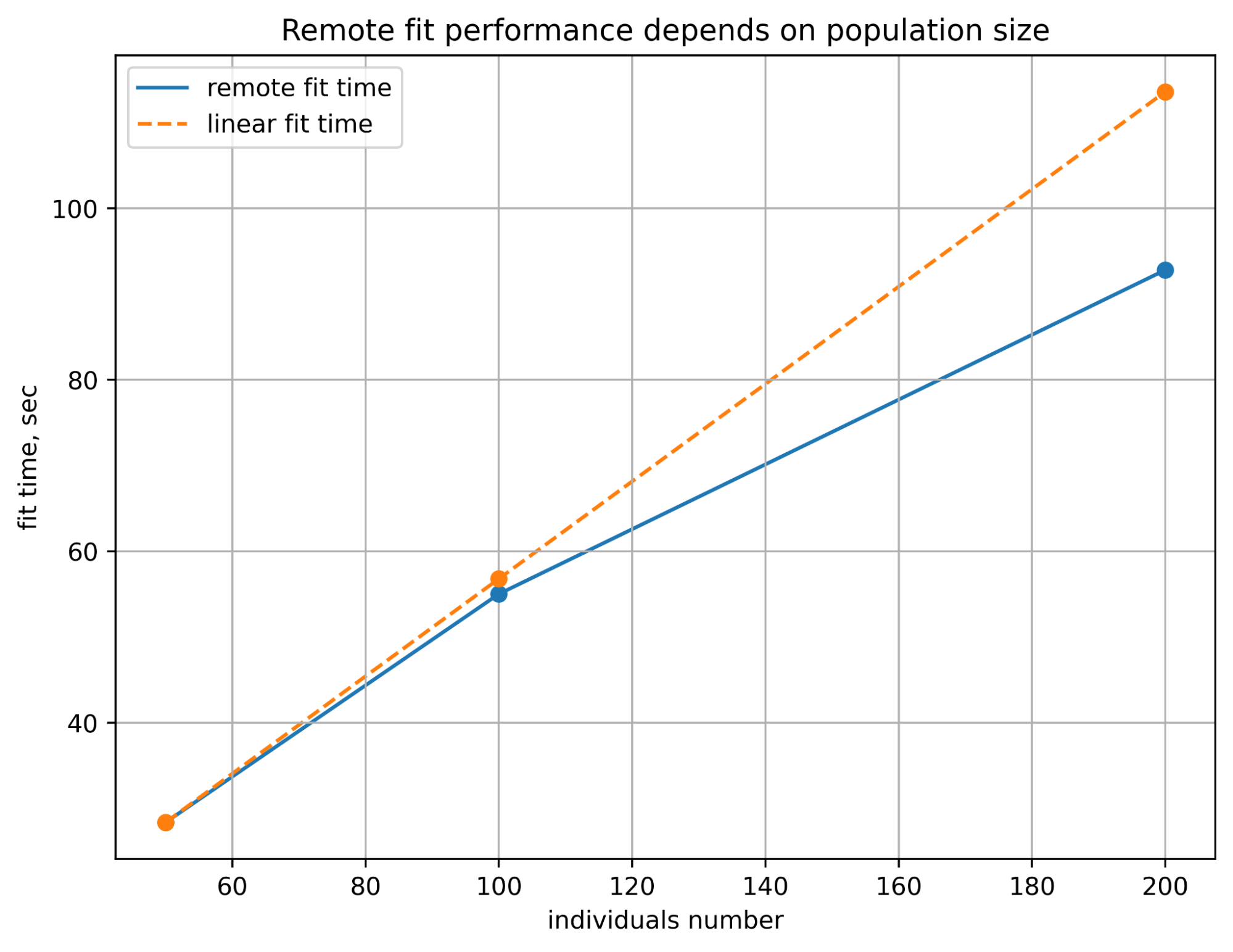}
         \caption{Without limits (fast calculations)}
         \label{fig_remote_training}
     \end{subfigure}
     \begin{subfigure}[b]{0.48\textwidth}
         \centering
         \includegraphics[width=\linewidth]{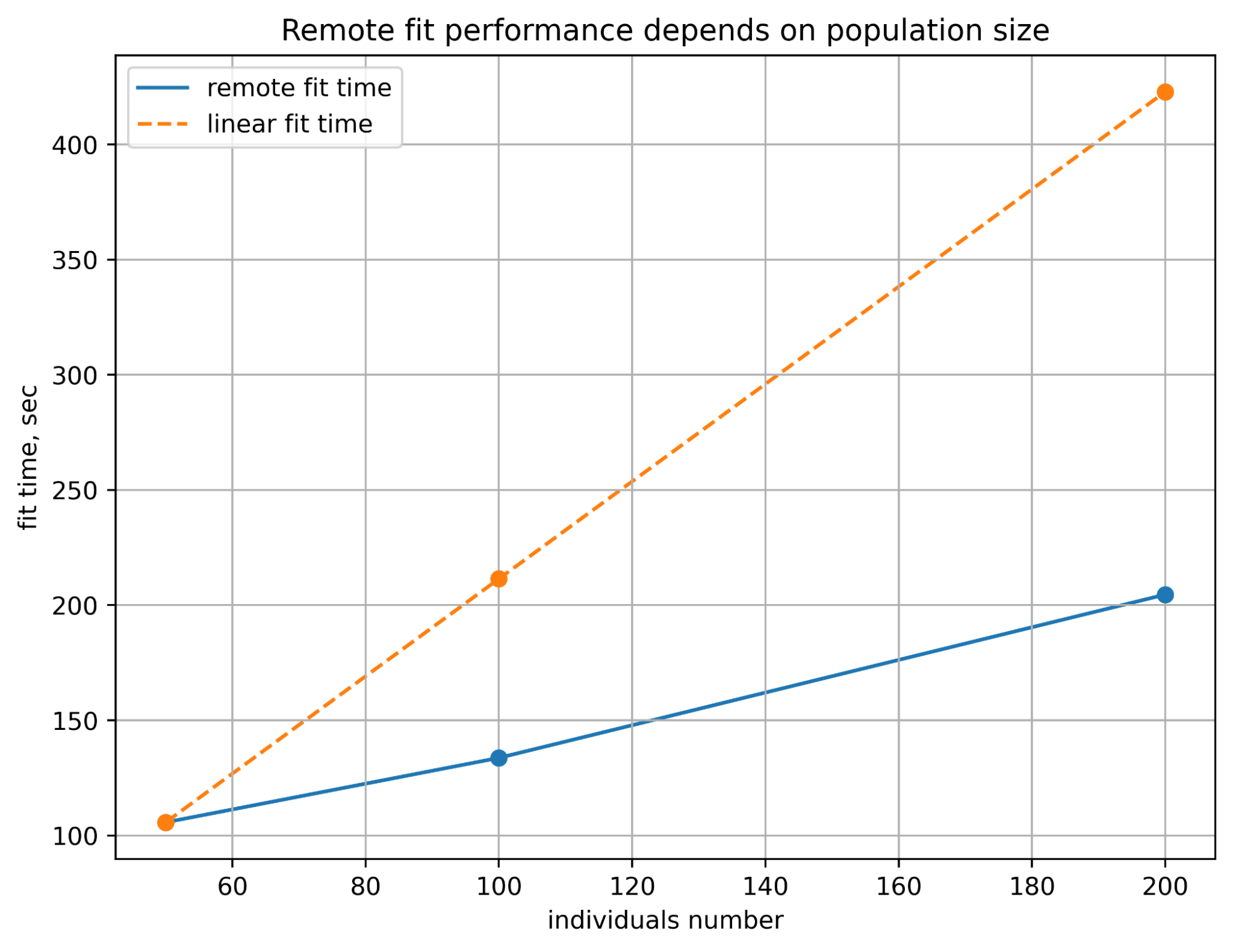}
         \caption{CPU limit = 0.2 core (heavy calculations emulation)}
         \label{fig_remote_heavy_training}
     \end{subfigure}
        \caption{The dependence of the total fit time on individual numbers in different computational setups. The orange line represents linear acceleration; the blue line represents the observed values of
        fit time.}
       \label{fig_all_remote_trainings}
\end{figure*}

Figure~\ref{fig_remote_training} presents training time depending on population size without results fetching. The evaluation of each individual has no CPU and system memory limits. Also, we have drawn a linear fit time as reference line. The initial point for this line is the time for the population size of 50. \ed{We found that training time increases almost linearly with the increase in population.}

\begin{figure*}[ht!] 
    \centering
    \includegraphics[width=15cm]{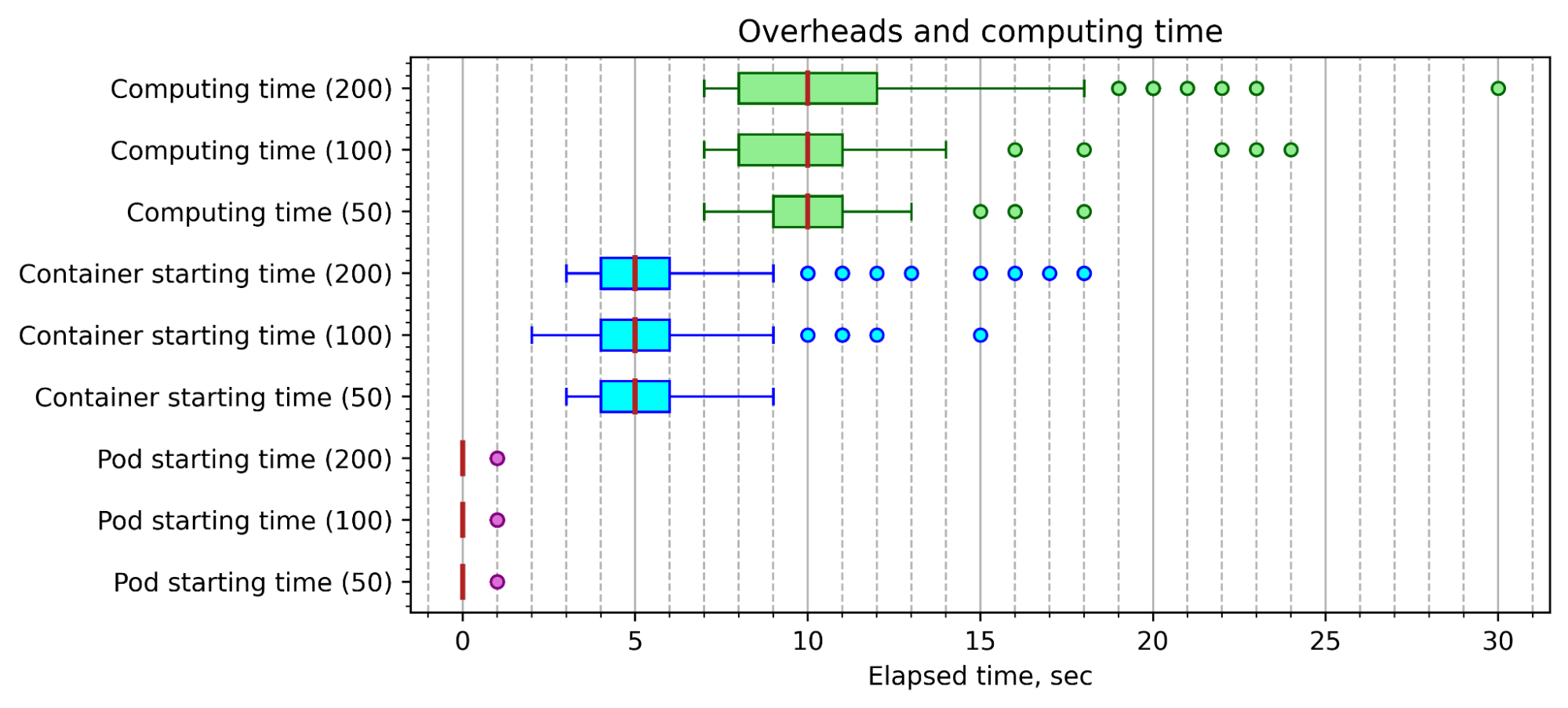}
    \caption{Overheads and computing time during experiments with remote infrastructure}
    \label{fig_time_consumption}
\end{figure*}

To explain the near-linear time growth, \ed{we can consider the operations that took the most time during the computing and the overheads they have}. Since requests overheads are nearly 0 seconds, they are not presented in Figure~\ref{fig_time_consumption}.

\begin{figure}[ht!] 
    \centering
    \includegraphics[width=8.5cm]{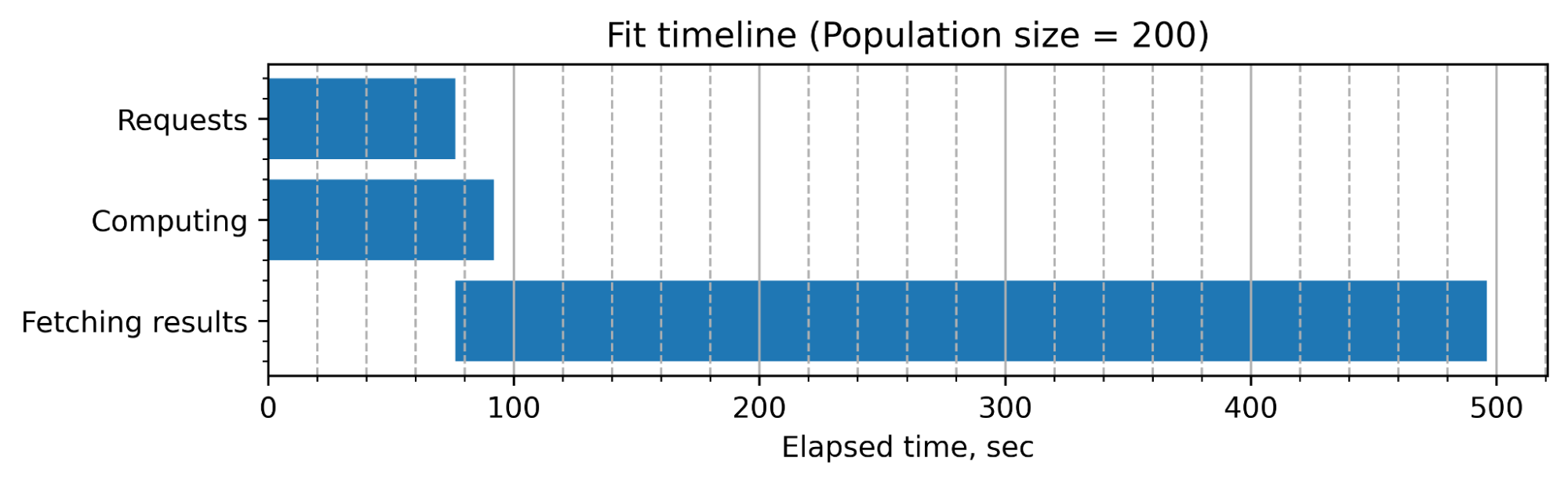}
    \caption{The explanation of remote training timeline with remote infrastructure.}
    \label{fig_remote_timeline}
\end{figure}

Figure~\ref{fig_time_consumption} shows that computing time and overheads for each individual are the same and are independent of population size. It means that the performance bottleneck is not on the cluster side. The fit stages timeline (\ed{including} results fetching) \ed{is} presented in Figure~\ref{fig_remote_timeline}.

This timeline shows that the most time is spent on results fetching. Result fetching consists of zip-file downloading over the network, unpacking and then deserializing the model. Since the results are fetched concurrently, a large number of individuals lead to a high network and drive load on the local machine.

Moreover, even though the overhead for the request to run one individual is less than 1 sec, a large number of requests also consume a significant amount of time. The time range between the last request and the last completed individual is less than 20 seconds, and it falls within the 75-percentile of computing time. It means that the computing cluster is underutilized because it is not running the individuals in parallel as well as expected because it is waiting for requests from the client.

To sum up, it is not reasonable to use remote training if we have lightweight and fast computations. Overheads in the form of requests and results fetching will be significantly larger than the payload. To prove this assumption, we have repeated the same experiment, but we have artificially introduced CPU limiting for each individual (up to 0.2 CPU core) to emulate "heavy" computing. The results for heavy-weight tasks are presented in Figure~\ref{fig_remote_heavy_training}.

We can conclude that remote computing provides a significant speedup for expensive computations. However, the overhead for small datasets should be taken into account.

\section{\ans{Conclusions and Discussions}}
\label{sec_conc}

In the paper, we propose a modular approach that improves the efficiency of evolutionary AutoML in a heterogeneous environment. \ans{The proposed approach differs from existing solutions since it can be configured for automated machine learning in various computational environments}. It makes it possible to parallelize and distribute the computational tasks across hybrid and/or remote computational systems. Also, caching algorithms are implemented to increase the optimization performance for composite pipelines.

 The AutoML-based experimental setup consisted of (1) the estimation of parallel speedup for a different number of processes; (2) an analysis of the efficiency of the cache; (3) an analysis of the GPU computations efficiency; (4) optimisation runs with remote infrastructure involved. The experiments confirm the proposed approach's efficiency. It allows achieving significant improvements in the number of evaluated individuals and \ed{in the fitness} function.

There are several ways to improve remote computing performance aimed at different bottlenecks that can be used separately or combined:
\begin{enumerate}
    \item Efficient cluster resources utilization requires a custom scheduler and additional plugins for batch workload such as Volcano\footnote{\url{https://volcano.sh}}. 
    \item Refuse to request to run each individual. Better to use one request to run a batch of individuals. This way, the number of requests will be reduced to one independent request for the all population. Also, we can apply speculative computing mode when the number of rest individuals is small;
    \item Provide a cluster file system mount \ed{on} the local machine. This will reduce the number of requests for downloading results, and the client will also skip zip file unpacking. Instead, the client will read the results from the mounted file system. If it is impossible, then we have to implement not only batch run requests but also batch download requests;
    \item Perform model validation using remote infrastructure too. This way, we \ed{also} have to provide a validation dataset to the remote system. Remote computing will validate individuals and save the score. It will make it unnecessary to fetch the trained models, and the calculated score will be enough for further decisions;
    \item Heterogeneous environment. We can use a heterogeneous environment not only on the cluster layer but on the client-server layer. For example, the client can perform lightweight calculations locally, heavy calculations at the same time will be sent to the cluster, and the heaviest calculations may be sent to the most powerful cluster nodes (e.g. special GPU nodes).
\end{enumerate}

Another direction of improvement is the support of large dataset processing. It can be based on the implementation of the distributed evaluation of different folds of the data set. The caching system \ed{can also} be implemented in a distributed way.

\section{\ans{Code and Data Availability}}
\label{sec_code}

The software implementation of all described methods and algorithms is available in the open repository \url{https://github.com/ITMO-NSS-team/fedot-performance-improvement-benchmark}.




\bibliography{sample}

\begin{thebibliography}{}

\bibitem[Bischl et~al., 2017]{bischl2017openml}
Bischl, B., Casalicchio, G., Feurer, M., Hutter, F., Lang, M., Mantovani,
  R.~G., van Rijn, J.~N., and Vanschoren, J. (2017).
\newblock Openml benchmarking suites and the openml100.
\newblock {\em stat}, 1050:11.

\bibitem[Cohen-Shapira and Rokach, 2022]{cohen2022learning}
Cohen-Shapira, N. and Rokach, L. (2022).
\newblock Learning dataset representation for automatic machine learning
  algorithm selection.
\newblock {\em Knowledge and Information Systems}, 64(10):2599--2635.

\bibitem[Erickson et~al., 2020]{erickson2020autogluon}
Erickson, N., Mueller, J., Shirkov, A., Zhang, H., Larroy, P., Li, M., and
  Smola, A. (2020).
\newblock Autogluon-tabular: Robust and accurate automl for structured data.
\newblock {\em arXiv preprint arXiv:2003.06505}.

\bibitem[Feurer et~al., 2020]{feurer2020auto}
Feurer, M., Eggensperger, K., Falkner, S., Lindauer, M., and Hutter, F. (2020).
\newblock Auto-sklearn 2.0: The next generation.
\newblock {\em arXiv preprint arXiv:2007.04074}.

\bibitem[Gao et~al., 2016]{GaoY2016}
Gao, Y., Sun, Y., and Wu, J. (2016).
\newblock Difference-genetic co-evolutionary algorithm for nonlinear mixed
  integer programming problems.
\newblock {\em J. Nonlinear Sci. Appl}, 9:1261--1284.

\bibitem[Gijsbers et~al., 2018]{gijsbers2018layered}
Gijsbers, P., Vanschoren, J., and Olson, R.~S. (2018).
\newblock Layered tpot: Speeding up tree-based pipeline optimization.
\newblock {\em arXiv preprint arXiv:1801.06007}.

\bibitem[Guerv{\'o}s et~al., 2019]{guervos2019improving}
Guerv{\'o}s, J.-J.~M., Laredo, J. L.~J., Castillo, P.~A., Valdez, M.~G., and
  Rojas-Galeano, S. (2019).
\newblock Improving the algorithmic efficiency and performance of channel-based
  evolutionary algorithms.
\newblock In {\em Proceedings of the Genetic and Evolutionary Computation
  Conference Companion}, pages 320--321.

\bibitem[He et~al., 2021]{he2021automl}
He, X., Zhao, K., and Chu, X. (2021).
\newblock Automl: A survey of the state-of-the-art.
\newblock {\em Knowledge-Based Systems}, 212:106622.

\bibitem[Kalyuzhnaya et~al., 2020]{kalyuzhnaya2020towards}
Kalyuzhnaya, A.~V., Nikitin, N.~O., Hvatov, A., Maslyaev, M., Yachmenkov, M.,
  and Boukhanovsky, A. (2020).
\newblock Towards generative design of computationally efficient mathematical
  models with evolutionary learning.
\newblock {\em Entropy}, 23(1):28.

\bibitem[Karatsiolis and Schizas, 2014]{karatsiolis2014implementing}
Karatsiolis, S. and Schizas, C.~N. (2014).
\newblock Implementing a software cache for genetic programming algorithms for
  reducing execution time.
\newblock In {\em IJCCI (ECTA)}, pages 259--265.

\bibitem[Kim, 2009]{kim2009effective}
Kim, K.~J. (2009).
\newblock {\em An effective distributed caching strategy for partial result
  reuse for discrete optimization problems}.
\newblock Dartmouth College.

\bibitem[K{\l}osko et~al., 2022]{klosko2022high}
K{\l}osko, J., Benecki, M., Wcis{\l}o, G., Dajda, J., and Turek, W. (2022).
\newblock High performance evolutionary computation with tensor-based
  acceleration.
\newblock In {\em Proceedings of the Genetic and Evolutionary Computation
  Conference}, pages 805--813.

\bibitem[Kratica, 1999]{kratica1999improving}
Kratica, J. (1999).
\newblock Improving performances of the genetic algorithm by caching.
\newblock {\em Computing and Informatics}, 18(3):271--283.

\bibitem[Le et~al., 2020]{le2020scaling}
Le, T.~T., Fu, W., and Moore, J.~H. (2020).
\newblock Scaling tree-based automated machine learning to biomedical big data
  with a feature set selector.
\newblock {\em Bioinformatics}, 36(1):250--256.

\bibitem[LeDell and Poirier, 2020]{ledell2020h2o}
LeDell, E. and Poirier, S. (2020).
\newblock H2o automl: Scalable automatic machine learning.
\newblock In {\em Proceedings of the AutoML Workshop at ICML}, volume 2020.

\bibitem[Li et~al., 2021]{li2021automl}
Li, Y., Wang, Z., Ding, B., and Zhang, C. (2021).
\newblock Automl: A perspective where industry meets academy.
\newblock In {\em Proceedings of the 27th ACM SIGKDD Conference on Knowledge
  Discovery \& Data Mining}, pages 4048--4049.

\bibitem[Majidi et~al., 2022]{majidi2022empirical}
Majidi, F., Openja, M., Khomh, F., and Li, H. (2022).
\newblock An empirical study on the usage of automated machine learning tools.
\newblock {\em arXiv preprint arXiv:2208.13116}.

\bibitem[Moritz et~al., 2018]{moritz2018ray}
Moritz, P., Nishihara, R., Wang, S., Tumanov, A., Liaw, R., Liang, E., Elibol,
  M., Yang, Z., Paul, W., Jordan, M.~I., et~al. (2018).
\newblock Ray: A distributed framework for emerging $\{$AI$\}$ applications.
\newblock In {\em 13th USENIX Symposium on Operating Systems Design and
  Implementation (OSDI 18)}, pages 561--577.

\bibitem[Nikitin et~al., 2020]{nikitin2020structural}
Nikitin, N.~O., Polonskaia, I.~S., Vychuzhanin, P., Barabanova, I.~V., and
  Kalyuzhnaya, A.~V. (2020).
\newblock Structural evolutionary learning for composite classification models.
\newblock {\em Procedia computer science}, 178:414--423.

\bibitem[Nikitin et~al., 2021]{nikitin2021automated}
Nikitin, N.~O., Vychuzhanin, P., Sarafanov, M., Polonskaia, I.~S., Revin, I.,
  Barabanova, I.~V., Maximov, G., Kalyuzhnaya, A.~V., and Boukhanovsky, A.
  (2021).
\newblock Automated evolutionary approach for the design of composite machine
  learning pipelines.
\newblock {\em Future Generation Computer Systems}.

\bibitem[Nikitin et~al., 2022]{nikitin2022automated}
Nikitin, N.~O., Vychuzhanin, P., Sarafanov, M., Polonskaia, I.~S., Revin, I.,
  Barabanova, I.~V., Maximov, G., Kalyuzhnaya, A.~V., and Boukhanovsky, A.
  (2022).
\newblock Automated evolutionary approach for the design of composite machine
  learning pipelines.
\newblock {\em Future Generation Computer Systems}, 127:109--125.

\bibitem[Olson and Moore, 2016]{olson2016tpot}
Olson, R.~S. and Moore, J.~H. (2016).
\newblock Tpot: A tree-based pipeline optimization tool for automating machine
  learning.
\newblock In {\em Workshop on automatic machine learning}, pages 66--74. PMLR.

\bibitem[Orzechowski and Moore, 2019a]{orzechowski2019mining}
Orzechowski, P. and Moore, a.~H. (2019a).
\newblock Mining a massive rna-seq dataset with biclustering: are evolutionary
  approaches ready for big data?
\newblock In {\em Proceedings of the Genetic and Evolutionary Computation
  Conference Companion}, pages 304--305.

\bibitem[Orzechowski and Moore, 2019b]{orzechowski2019strategies}
Orzechowski, P. and Moore, J.~H. (2019b).
\newblock Strategies for improving performance of evolutionary biclustering
  algorithm ebic.
\newblock In {\em Proceedings of the Genetic and Evolutionary Computation
  Conference Companion}, pages 185--186.

\bibitem[Packard et~al., 2019]{packard2019open}
Packard, N., Bedau, M.~A., Channon, A., Ikegami, T., Rasmussen, S., Stanley,
  K., and Taylor, T. (2019).
\newblock Open-ended evolution and open-endedness: Editorial introduction to
  the open-ended evolution i special issue.

\bibitem[Parmentier et~al., 2019]{parmentier2019tpot}
Parmentier, L., Nicol, O., Jourdan, L., and Kessaci, M.-E. (2019).
\newblock Tpot-sh: A faster optimization algorithm to solve the automl problem
  on large datasets.
\newblock In {\em 2019 IEEE 31st International Conference on Tools with
  Artificial Intelligence (ICTAI)}, pages 471--478. IEEE.

\bibitem[Pedregosa et~al., 2011]{scikit-learn}
Pedregosa, F., Varoquaux, G., Gramfort, A., Michel, V., Thirion, B., Grisel,
  O., Blondel, M., Prettenhofer, P., Weiss, R., Dubourg, V., Vanderplas, J.,
  Passos, A., Cournapeau, D., Brucher, M., Perrot, M., and Duchesnay, E.
  (2011).
\newblock Scikit-learn: Machine learning in {P}ython.
\newblock {\em Journal of Machine Learning Research}, 12:2825--2830.

\bibitem[Polonskaia et~al., 2021]{polonskaia2021multi}
Polonskaia, I.~S., Nikitin, N.~O., Revin, I., Vychuzhanin, P., and Kalyuzhnaya,
  A.~V. (2021).
\newblock Multi-objective evolutionary design of composite data-driven models.
\newblock In {\em 2021 IEEE Congress on Evolutionary Computation (CEC)}, pages
  926--933.

\bibitem[Przewozniczek and Komarnicki, 2021]{przewozniczek2021fitness}
Przewozniczek, M.~W. and Komarnicki, M.~M. (2021).
\newblock Fitness caching-from a minor mechanism to major consequences in
  modern evolutionary computation.
\newblock In {\em 2021 IEEE Congress on Evolutionary Computation (CEC)}, pages
  1785--1791. IEEE.

\bibitem[Santos and Santos~Jr, 2001]{santos2001effective}
Santos, E.~E. and Santos~Jr, E. (2001).
\newblock Effective and efficient caching in genetic algorithms.
\newblock {\em International Journal on Artificial Intelligence Tools},
  10(01n02):273--301.

\bibitem[Santu et~al., 2020]{santu2020automl}
Santu, S. K.~K., Hassan, M., Smith, M.~J., Xu, L., Zhai, C., Veeramachaneni,
  K., et~al. (2020).
\newblock Automl to date and beyond: Challenges and opportunities.
\newblock {\em arXiv preprint arXiv:2010.10777}.

\bibitem[Silva et~al., 2010]{DaSilva2010}
Silva, F. J. M.~D., Pérez, J. M.~S., Pulido, J. A.~G., and Rodríguez, M.~A.
  (2010).
\newblock Parallel alineaga: An island parallel evolutionary algorithm for
  multiple sequence alignment.
\newblock {\em Proceedings of the 2010 International Conference of Soft
  Computing and Pattern Recognition, SoCPaR 2010}, pages 279--284.

\bibitem[Singh and Joshi, 2022]{singh2022automated}
Singh, V.~K. and Joshi, K. (2022).
\newblock Automated machine learning (automl): an overview of opportunities for
  application and research.
\newblock {\em Journal of Information Technology Case and Application
  Research}, pages 1--11.

\bibitem[Team, 2018]{rapids}
Team, R.~D. (2018).
\newblock {\em RAPIDS: Collection of Libraries for End to End GPU Data
  Science}.

\bibitem[Tornede et~al., 2021]{tornede2021towards}
Tornede, T., Tornede, A., Hanselle, J., Wever, M., Mohr, F., and
  H{\"u}llermeier, E. (2021).
\newblock Towards green automated machine learning: Status quo and future
  directions.
\newblock {\em arXiv preprint arXiv:2111.05850}.

\bibitem[Vakhrushev et~al., 2021]{vakhrushev2021lightautoml}
Vakhrushev, A., Ryzhkov, A., Savchenko, M., Simakov, D., Damdinov, R., and
  Tuzhilin, A. (2021).
\newblock Lightautoml: Automl solution for a large financial services
  ecosystem.
\newblock {\em arXiv preprint arXiv:2109.01528}.

\bibitem[Visheratin et~al., 2020]{visheratin2020peregreen}
Visheratin, A., Struckov, A., Yufa, S., Muratov, A., Nasonov, D., Butakov, N.,
  Kuznetsov, Y., and May, M. (2020).
\newblock Peregreen--modular database for efficient storage of historical time
  series in cloud environments.
\newblock In {\em 2020 USENIX Annual Technical Conference (USENIX ATC 20)},
  pages 589--601.

\bibitem[Zaharia et~al., 2018]{zaharia2018accelerating}
Zaharia, M., Chen, A., Davidson, A., Ghodsi, A., Hong, S.~A., Konwinski, A.,
  Murching, S., Nykodym, T., Ogilvie, P., Parkhe, M., et~al. (2018).
\newblock Accelerating the machine learning lifecycle with mlflow.
\newblock {\em IEEE Data Eng. Bull.}, 41(4):39--45.

\bibitem[Z{\"o}ller and Huber, 2019]{zoller2019benchmark}
Z{\"o}ller, M.-A. and Huber, M.~F. (2019).
\newblock Benchmark and survey of automated machine learning frameworks.
\newblock {\em arXiv preprint arXiv:1904.12054}.

\end{thebibliography}

\end{document}